\documentclass[10pt,aps,pre,superscriptaddress,notitlepage,twocolumn]{revtex4-1}

\usepackage{amsfonts,amssymb,amsmath,latexsym,epsfig,wasysym} 
\usepackage[sort&compress]{natbib}
\usepackage{color}
\definecolor{bluecolor}{rgb}{0,0.,1.}

\definecolor{redcolor}{rgb}{.7,0.,0.}

\usepackage{graphicx}
\usepackage{dcolumn}
\usepackage{bm}
\usepackage{hyperref}

\usepackage{lineno} 

\newcommand{\dd}{\mathrm{d}}

\usepackage{mathtools}
\def\multiset#1#2{\ensuremath{\left(\kern-.3em\left(\genfrac{}{}{0pt}{}{#1}{#2}\right)\kern-.3em\right)}}

\usepackage{amsmath}

\definecolor{bluecolor}{rgb}{0,0.,1.}

\definecolor{redcolor}{rgb}{.7,0.,0.}

\begin{document}


\title{A network approach to topic models}

\author{Martin Gerlach}
\affiliation{ Department of Chemical and Biological Engineering, Northwestern University, Evanston, IL 60208, USA}
\affiliation{Max Planck Institute for the Physics of Complex Systems, D-01187 Dresden, Germany}

\author{Tiago P. Peixoto}
\affiliation{Department of Mathematical Sciences and Centre for Networks
and Collective Behaviour, University of Bath, Claverton Down, Bath BA2
7AY, United Kingdom}
\affiliation{ISI Foundation, Via Alassio 11/c, 10126 Torino, Italy}

\author{Eduardo G. Altmann}
\affiliation{Max Planck Institute for the Physics of Complex Systems, D-01187 Dresden, Germany}
\affiliation{School of Mathematics and Statistics, University of Sydney, 2006 NSW, Australia}



\begin{abstract}
One of the main computational and scientific challenges in the modern
age is to extract useful information from unstructured texts.  Topic models are one popular
machine-learning approach which infers the latent topical structure of a
collection of documents.  Despite their success --- in particular of its
most widely used variant called Latent Dirichlet Allocation (LDA) ---
and numerous applications in sociology, history, and linguistics, topic
models are known to suffer from severe conceptual and practical
problems, e.g. a lack of justification for the Bayesian priors,
discrepancies with statistical properties of real texts, and the
inability to properly choose the number of topics.
Here we obtain a fresh view on the problem of identifying topical structures by relating it to the problem of finding communities in complex networks. 
This is achieved by representing text corpora as bipartite networks of documents and words. 
By adapting existing community-detection methods -- using a stochastic block model (SBM) with non-parametric priors -- we obtain a more versatile and principled framework for topic modeling (e.g., it automatically detects the number of topics and hierarchically clusters both the words and documents). 
The analysis of artificial and real corpora demonstrates that our SBM approach leads to better topic models than LDA in terms of statistical model selection.  
More importantly, our work shows how to formally relate methods from community detection and topic modeling, opening the possibility of cross-fertilization between these two fields.
\end{abstract}

\maketitle


\section{Introduction}
\label{sec.intro}

The accelerating rate of digitization of information increases
the importance and number of problems which require automatic
organization and classification of written text.  Topic models~\cite{blei.2012}  are a
flexible and widely used tool which identifies
semantically related documents through the topics they address.  
These methods originated in
machine learning %
and were largely based on heuristic
approaches such as singular value decomposition in latent semantic
indexing (LSI)~\cite{deerwester.1990} in which one optimizes an arbitrarily chosen quality function.
Only a more statistically
principled approach, based on the formulation of probabilistic
generative models~\cite{ghahramani.2015}, allowed for a deeper
theoretical foundation within the framework of Bayesian statistical
inference. %
This, in turn, lead to a series of
key developments, in particular probabilistic latent semantic indexing
(pLSI)~\cite{hofmann.1999} and latent Dirichlet allocation
(LDA)~\cite{blei.2003,griffiths.2004}.  The latter established itself as
the state-of-the-art method in topic modeling and has been widely used
not only for recommendation and classification~\cite{manning.book2008}
but also bibliometrical~\cite{boyack.2011},
psychological~\cite{mcnamara.2011}, and political~\cite{grimmer.2013}
analysis.  Beyond the scope of natural language, LDA has also been
applied in biology~\cite{liu.2010} (developed
independently in this context~\cite{pritchard.2000}), or image
processing~\cite{fei.2005}.

However, despite its success and overwhelming popularity, LDA is known
to suffer from fundamental flaws in the way it represents text. In
particular, it lacks an intrinsic methodology to choose the number of
topics, and contains a large number of free parameters that can cause
overfitting.  Furthermore, there is no 
justification for the
use of the Dirichlet prior in the model formulation besides mathematical convenience.
This choice restricts the types of topic mixtures and is not designed to be compatible with well-known properties of real
text~\cite{altmann.book2016}, such as Zipf's law~\cite{zipf.1936} for
the frequency of words.
More recently, consistency problems have also
been identified with respect to how planted structures in artificial
corpora can be recovered with LDA~\cite{lancichinetti.2015}.  A
substantial part of the research in topic models focuses on creating
more sophisticated and realistic versions of LDA that account for,
e.g., syntax~\cite{griffiths.2005}, correlations between
topics~\cite{li.2006}, meta-information (such as
authors)~\cite{rosen.2004}, or burstiness~\cite{doyle.2009}.  
Other
approaches consist of post-inference fitting of the number of
topics~\cite{zhao_heuristic_2015} or the
hyperparameters~\cite{wallach.2009a}, or the formulation
of nonparametric hierarchical
extensions~\cite{teh.2006,blei.2010,Paisley2015}.  
In particular, models based on the Pitman-Yor~\cite{Sudderth2009,Sato2010,Buntine2014}
or the negative binomial process have tried to address the issue of Zipf's law~\cite{Broderick2015} yielding useful generalizations of the simplistic Dirichlet prior~\cite{Zhou2015}. 
While all these
approaches lead to demonstrable improvements, they do not provide
satisfying solutions to the aforementioned issues because they either
share the limitations due to the choice of Dirichlet priors, introduce
idiosyncratic structures to the model, or rely on heuristic approaches
in the optimization of the free parameters.

A similar evolution from heuristic approaches to probabilistic models is occurring in the field of complex networks, in particular in the problem of community detection~\cite{fortunato.2010}. 
Topic models and community-detection methods have been developed largely independently from each other with only a few papers pointing to their conceptual similarities~\cite{airoldi.2007,ball.2011,lancichinetti.2015}.
The idea of community
detection is to find large-scale structure, i.e. the identification of
groups of nodes with similar connectivity patterns~\cite{fortunato.2010}.  This is motivated
by the fact that these groups describe the heterogeneous nonrandom structure of the network and may correspond to functional
units, giving potential insights on the generative mechanisms behind the
network formation. While there is a variety of
different approaches to community detection, most
methods are heuristic and optimize a
quality function, the most popular being
modularity~\cite{newman.2004}.  Modularity suffers from severe conceptual
deficiencies, such as its inability to assess statistical significance
leading to detection of groups in completely random
networks~\cite{guimera.2004}, or its incapacity in finding groups below
a given size~\cite{lancichinetti.2011}.  Methods like
modularity maximization are analogous to the pre-pLSI heuristic
approaches to topic models, sharing with them many conceptual and
practical deficiencies. In an effort to quench these problems, many
researchers moved to probabilistic inference approaches, most notably
those based on stochastic block models
(SBM)~\cite{holland.1983,airoldi.2007,karrer.2011}, mirroring the same
trend that occurred in topic modeling.

\begin{figure}[tb]
\centerline{\includegraphics[width=1\columnwidth]{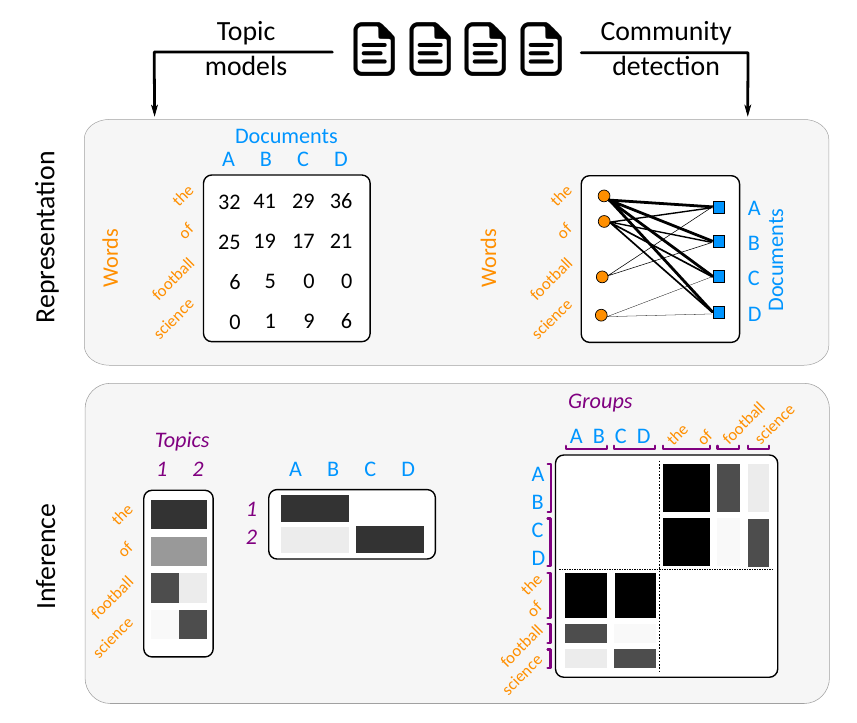}}
\caption{Two approaches to extract information from collections of texts.  Topic models
  represent the texts as a document-word matrix (how often each word appears in each
  document) which is then written as a product of two matrices of smaller dimensions with
  the help of the latent variable topic.  The approach we propose here represents texts as a
  network and infers communities in this network.  The nodes consists of
  documents and words and the  strength of the edge between them is given by the number of
  occurrences of the word in the document, yielding a bipartite multigraph that is
  equivalent to the word-document matrix used in topic models. \label{fig.schematic}}

\end{figure}

In this paper we propose and apply a unified framework to the fields of topic modeling and community detection. 
As illustrated in Fig.~\ref{fig.schematic}, by representing the word-document matrix as a bipartite network the problem of inferring topics becomes a problem of
inferring communities.
Topic models and community-detection methods have been previously discussed as being part of mixed-membership models~\cite{Airoldi2014}. 
However, this has remained a conceptual connection~\cite{lancichinetti.2015} and in practice the two approaches are used to address different problems~\cite{airoldi.2007}; the occurrence of words within and the links/citations between documents, respectively. 
In contrast, here we develop a formal correspondence that builds on the mathematical equivalence between  pLSI of texts and SBMs of networks~\cite{ball.2011}  and that we use to adapt community-detection methods to perform topic modeling.
In particular, we derive a nonparametric Bayesian parametrization of
pLSI --- adapted from a hierarchical stochastic block model
(hSBM)~\cite{peixoto.2014a,peixoto.2015,peixoto_nonparametric_2017} ---
that makes fewer assumptions about the underlying structure of the data.
As a consequence, it better matches the statistical properties of
real texts and solves many of the intrinsic limitations of LDA.
For example, we demonstrate the limitations induced by the Dirichlet priors by showing that LDA fails to infer topical structures that deviate from the Dirichlet assumption.
We show that our model infers correctly such structures and thus leads to a better topic model than Dirichlet-based methods (such as LDA) in the terms of model selection not only in various real corpora but even in artificial corpora generated from LDA itself.
Additionally, our nonparametric approach uncovers topical structures on many scales of resolution, automatically determines the
number of topics together with the word classification, and its
symmetric formulation allows the documents themselves to be clustered
into hierarchical categories.

The goal of our manuscript is to introduce a unified approach to topic modeling and community detection, showing how ideas and methods can be transported between these two classes of problems. 
The benefit of this unified approach is illustrated by the derivation of an alternative to Dirichlet-based topic models, which is more principled in its theoretical foundation (making fewer assumption about the data) and superior in practice according to model selection criteria.


\section{Results}
\subsection{Community Detection for Topic Modeling}
\label{sec.theory}
\begin{figure}[tb]
\centerline{\includegraphics[width=1\columnwidth]{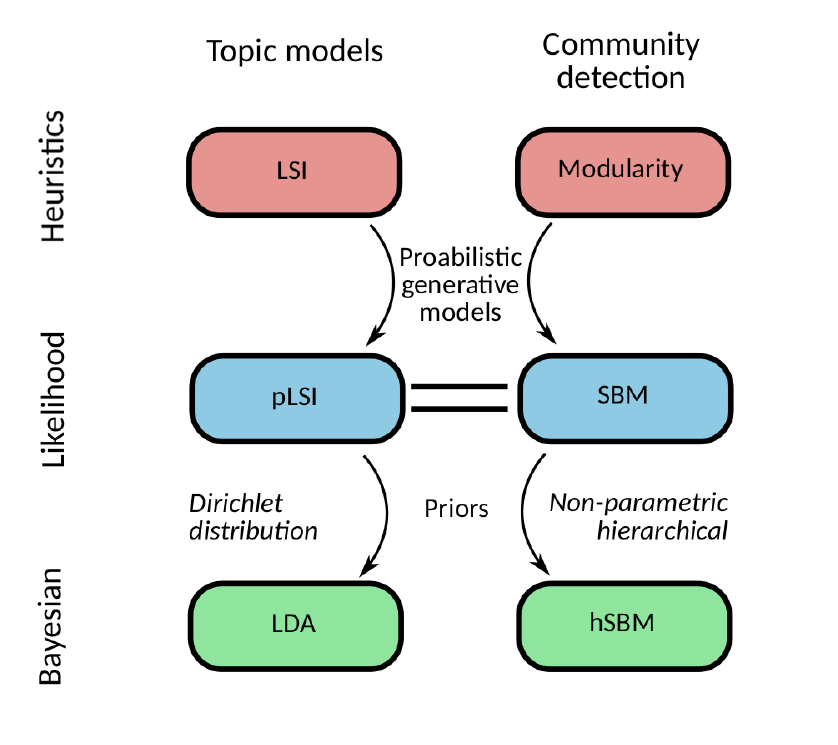}}
\caption{Parallelism between topic models and community detection
methods. The probabilistic latent semantic indexing (pLSI) and stochastic
block models (SBM) are mathematically equivalent and therefore methods
from community detection (e.g., the hSBM we propose in this manuscript)
can be used as alternatives to traditional topic models (e.g., LDA).
\label{fig.ldasbm.prior}}

\end{figure}

In this section we expose the connection between topic modeling and
community detection, as illustrated in Fig.~\ref{fig.ldasbm.prior}.  We
first revisit how a Bayesian formulation of pLSI assuming Dirichlet
priors leads to LDA and how the former can be re-interpreted as a
mixed membership SBM.  
We then use the latter to derive a more
principled approach to topic modeling using nonparametric and
hierarchical priors.

\subsubsection{Topic models: pLSI and LDA}
\label{sec.theory.plsi}

PLSI is a model that generates a corpus composed of $D$ documents, where
each document $d$ has $k_d$ words~\cite{hofmann.1999}. The placement of the words in the
documents is done based on the assignment of topic mixtures to both
document and words, from a total of $K$ topics. More specifically, one
iterates through all $D$ documents, and for each document $d$ one
samples $k_d \sim \operatorname{Poi}(\eta_d)$ and for each word-token $l
\in \{1,k_d\}$, first a topic $r$ is chosen with probability
$\theta_{dr}$, and then a word $w$ is chosen from that topic with
probability $\phi_{rw}$. If $n_{dw}^r$ is the number of occurrences of
word $w$ of topic $r$ in document $d$ (summarized as $\bm{n}$), the probability of a corpus is
\begin{equation}
\label{eq.plsi}
P(\bm{n}|\bm{\eta}, \bm{\theta}, \bm{\phi}) =
\prod_d \eta_d^{k_d} e^{-\eta_d}\prod_{wr}\frac{(\phi_{rw}\theta_{dr})^{n_{dw}^r}}{n_{dw}^r!}.
\end{equation}
We denote matrices by bold-face symbols, e.g. $\bm{\theta} = \{ \theta_{dr} \}$ with $d=1,\ldots,D$ and $r=1,\dots,K$ where $\theta_{dr}$ is an individual entry, thus the notation $\bm{\theta}_{d} $ refers to the vector $\{ \theta_{dr}\}$ with fixed $d$ and $r=1,\ldots,K$.

For an unknown text, we could simply maximize Eq.~\eqref{eq.plsi} to obtain
the best parameters $\bm{\eta}$, $\bm{\theta}$, and $\bm{\phi}$ which
describe the topical structure of the corpus.  However, this approach
cannot be used directly to model textual data without a significant
danger of overfitting.  The model possesses a large number of parameters,
that grows as the number of documents, words, and topics is increased,
and hence a maximum likelihood estimate will invariably incorporate a
considerable amount of noise.  One solution to this problem is to employ
a Bayesian formulation, by proposing prior distributions to the
parameters, and integrating over them.  This is precisely what is done
in LDA~\cite{blei.2003,griffiths.2004}, where one chooses Dirichlet
priors $D_d(\bm{\theta}_d|\bm{\alpha}_d)$ and
$D_r(\bm{\phi}_r|\bm{\beta}_r)$ with hyperparameters $\bm{\alpha}$ and
$\bm{\beta}$ for the probabilities $\bm{\theta}$ and $\bm{\phi}$ above,
and one uses instead the marginal likelihood.
\begin{multline}\label{eq.marginalL}
  P(\bm{n}|\bm{\eta}, \bm{\beta}, \bm{\alpha}) \\
  \begin{aligned}
  \quad&=
  \int
  P(\bm{n}|\bm{\eta}, \bm{\theta}, \bm{\phi})\prod_dD_d(\bm{\theta}_d|\bm{\alpha}_d)\prod_rD_r(\bm{\phi}_r|\bm{\beta}_r)\;\dd \bm{\theta}\dd\bm{\phi},\\
  &=\prod_d \eta_d^{k_d} e^{-\eta_d}\prod_{wr}\frac{1}{n_{dw}^r!} \times \\
  &\qquad\prod_d\frac{\Gamma(\sum_r\alpha_{dr})}{\Gamma(k_d+\sum_r\alpha_{dr})}\prod_r\frac{\Gamma(\sum_wn_{dw}^r+\alpha_{dr})}{\Gamma(\alpha_{dr})} 
  \times \\
  &\qquad\prod_r\frac{\Gamma(\sum_w\beta_{rw})}{\Gamma(\sum_{dw}n_{dw}^r+\sum_w\beta_{rw})}\prod_w\frac{\Gamma(\sum_dn_{dw}^r+\beta_{rw})}{\Gamma(\beta_{rw})},
  \end{aligned}
\end{multline}
If one makes a noninformative choice, i.e. $\alpha_{dr}=1$ and
$\beta_{rw}=1$, inference using Eq.~\eqref{eq.marginalL} is
nonparametric and less susceptible to overfitting. In particular, one
can obtain the labeling of word-tokens into topics, $n_{dw}^r$,
conditioned only on the observed total frequencies of words in
documents, $\sum_rn_{dw}^r$, in addition to the number of topics $K$
itself, simply by maximizing or sampling from the posterior
distribution. The weakness of this approach rests in the fact that the
Dirichlet prior is a simplistic assumption about the data-generating
process: In its noninformative form, every mixture in the model --- both
of topics in each document as well as words into topics --- is assumed
to be equally likely, precluding the existence of any form of
higher-order structure. This limitation has prompted the widespread
practice of inferring using LDA in a parametric way, by maximizing the
likelihood with respect to the hyperparameters $\bm{\alpha}$ and
$\bm{\beta}$, which can improve the quality of fit in many cases. But
not only this undermines to a large extent the initial purpose of a
Bayesian approach --- as the number of hyperparameters still increases
with the number of documents, words and topics, and hence maximizing
over them reintroduces the danger of overfitting --- but also it does
not sufficiently addresses the original limitation of the Dirichlet
prior. Namely, regardless of the hyperparameter choice, the Dirichlet
distribution is \emph{unimodal}, meaning that it generates mixtures
which are either concentrated around the mean value, or spread away
uniformly from it towards pure components. This means that for any
choice of $\bm{\alpha}$ and $\bm{\beta}$ the whole corpus is
characterized by a single typical mixture of topics into documents, and
a single typical mixture of words into topics. This is an extreme level
of assumed homogeneity which stands in contradiction to a clustering
approach initially designed to capture heterogeneity.

In addition to the above, the use of nonparametric Dirichlet priors is
inconsistent with well-known universal statistical properties of real
texts; most notably the highly-skewed distribution of word frequencies,
which typically follows Zipf's law~\cite{zipf.1936}. In contrast, the
noninformative choice of the Dirichlet distribution with hyperparameters
$\beta_{rw}=1$ amounts to an expected \emph{uniform} frequency of words
in topics and documents. Although this disagreement can be addressed by
choosing appropriate values of $\beta_{rw}$, such an approach, as
already mentioned, runs contrary to nonparametric inference, and is
subject to overfitting.

In the following, we will show how the same original pLSI model can be
re-cast as a network model that completely removes the limitations
described above, and is capable of uncovering heterogeneity in the data
at multiple scales.


\subsubsection{Topic models and community detection: Equivalence between pLSI and SBM}
\label{sec.theory.plsisbm}

We show that pLSI is equivalent to a specific form of a mixed membership
SBM as proposed by Ball et al.~\cite{ball.2011}.

The SBM is a model that generates a network composed of $i=1,\ldots,N$
nodes with adjacency matrix $A_{ij}$, which we will assume without loss
of generality to correspond to a multigraph, i.e. $A_{ij}\in
\mathbb{N}$. The nodes are placed in a partition composed of $B$
overlapping groups, and the edges between nodes $i$ and $j$ are sampled
from a Poisson distribution with average
\begin{equation}
  \sum_{rs}\kappa_{ir}\omega_{rs}\kappa_{js},
\end{equation}
where $\omega_{rs}$ is the expected number of edges between group $r$
and group $s$, and $\kappa_{ir}$ is the probability that node $i$ is
sampled from group $r$. The likelihood to observe
$\bm{\mathcal{A}}=\{\mathcal{A}_{ij}^{rs}\}$, i.e. a particular
decomposition of $A_{ij}$ into labeled half-edges (i.e. edge endpoints)
such that $A_{ij}=\sum_{rs}\mathcal{A}_{ij}^{rs}$, can be written as
\begin{align}
\label{eq.sbm}
  P(\bm{\mathcal{A}}|\bm{\kappa},\bm{\omega}) &=
  \prod_{i<j}\prod_{rs}\frac{e^{-\kappa_{ir}\omega_{rs}\kappa_{is}}(\kappa_{ir}\omega_{rs}\kappa_{js})^{\mathcal{A}_{ij}^{rs}}}{\mathcal{A}_{ij}^{rs}!}\times\nonumber\\
  &\qquad\prod_i\prod_{rs}\frac{e^{-\kappa_{ir}\omega_{rs}\kappa_{is}/2}(\kappa_{is}\omega_{rs}\kappa_{is}/2)^{\mathcal{A}_{ii}^{rs}/2}}{\mathcal{A}_{ii}^{rs}/2!},
\end{align}
by exploiting the fact that the sum of Poisson variables is also
distributed according to a Poisson.

The connection to pLSI can now be made by rewriting the token
probabilities in Eq.~\eqref{eq.plsi} in a symmetric fashion as
\begin{equation}
  \phi_{rw}\theta_{dr} = \eta_w\theta_{dr}\phi'_{wr},
\end{equation}
where $\phi'_{wr} \equiv \phi_{rw} / \sum_s\phi_{sw}$ is the probability
that the word $w$ belongs to topic $r$, and $\eta_w \equiv
\sum_s\phi_{sw}$ is the overall propensity with which the word $w$ is chosen across all topics. 
In this manner, the likelihood of Eq.~\eqref{eq.plsi} can be
re-written as
\begin{equation}
\label{eq.plsi.2}
  P(\bm{n}| \bm{\eta}, \bm{\phi}', \bm{\theta}) = \prod_{dwr} \frac{e^{-\lambda_{dw}^r}(\lambda_{dw}^r)^{n_{dw}^r}}{n_{dw}^r!},
\end{equation}
with $\lambda_{dw}^r = \eta_d\eta_w\theta_{dr}\phi'_{wr}$.  If we choose
to view the counts $n_{dw}$ as the entries of the adjacency matrix of a
bipartite multigraph with documents and words as nodes, the
likelihood of Eq.~\eqref{eq.plsi.2} is equivalent to the
likelihood of Eq.~\eqref{eq.sbm} of the SBM, if we assume that each document
belongs to its own specific group, $\kappa_{ir}=\delta_{ir}$, with
$i=1,\dots,D$ for document-nodes, and by re-writing
$\lambda_{dw}^r=\omega_{dr}\kappa_{rw}$. Therefore, the SBM of
Eq.~\eqref{eq.sbm} is a generalization of pLSI that allows the words as
well as the documents to be clustered into groups, and includes it as a
special case when the documents are not clustered. 

In the symmetric setting of the SBM, we make no explicit distinction
between words and documents, both of which become nodes in different
partitions of a bipartite network.  We base our Bayesian formulation
that follows on this symmetric parametrization.


\subsubsection{Community detection and the hierarchical SBM}
\label{sec.theory.hsbm}
Taking advantage of the above connection between pLSI and SBM, we show
how the idea of hierarchical SBMs developed in
Refs.~\cite{peixoto.2014a,peixoto.2015,peixoto_nonparametric_2017} can
be extended such that they can be effectively used for the inference of
topical structure in texts.

Like pLSI, the SBM likelihood of Eq.~\eqref{eq.sbm} contains a large
number of parameters that grows with the number of groups, and therefore
cannot be used effectively without knowing the most appropriate
dimension of the model beforehand. Analogously to what is done in LDA,
this can be addressed by assuming noninformative priors for the
parameters $\bm{\kappa}$ and $\bm{\omega}$, and computing the marginal
likelihood (for an explicit expression see \textit{Supplementary Materials} Sec. 1.1)
\begin{equation}\label{eq.marginal}
  P(\bm{\mathcal{A}}|\bar\omega) = \int P(\bm{\mathcal{A}}|\bm{\kappa},\bm{\omega})P(\bm{\kappa})P(\bm{\omega}|\bar{\omega})\; \dd\bm{\kappa}\dd\bm{\omega},
\end{equation}
where $\bar\omega$ is a global parameter determining the overall density
of the network. This can be used to infer the labeled adjacency matrix
$\{\mathcal{A}_{ij}^{rs}\}$ as done in LDA, with the difference
that not only the words but also the documents would be clustered into
mixed categories.

However, at this stage the model still shares some disadvantages with
LDA. In particular, the noninformative priors make unrealistic
assumptions about the data, where the mixture between groups and the
distribution of nodes into groups is expected to be unstructured. Among
other problems, this leads to a practical obstacle, as this approach
possesses a ``resolution limit'' where at most $O(\sqrt{N})$ groups can
be inferred on a sparse network with $N$
nodes~\cite{peixoto_parsimonious_2013,peixoto_nonparametric_2017}.  In
the following we propose a qualitatively different approach to the
choice of priors by replacing the noninformative approach with deeper
Bayesian hierarchy of priors and hyperpriors, which are agnostic about the higher order properties of the data while maintaining the nonparametric nature of the approach. We begin by
re-formulating the above model as an equivalent
\emph{microcanonical} model~\cite{peixoto_nonparametric_2017} (for a proof see  \textit{Supplementary Materials} Sec. 1.2) 
such that we can write the marginal likelihood as the joint likelihood of the data and its discrete
parameters,
\begin{equation}
  P(\bm{\mathcal{A}}|\bar\omega) =  P(\bm{\mathcal{A}},\bm{k},\bm{e}|\bar\omega) = P(\bm{\mathcal{A}}|\bm{k},\bm{e})P(\bm{k}|\bm{e})P(\bm{e}|\bar\omega),
\end{equation}
with 
\begin{align}
  P(\bm{\mathcal{A}}|\bm{k},\bm{e}) &= \frac{\prod_{r<s}e_{rs}!\prod_re_{rr}!!\prod_{ir}k_i^r!}{\prod_{rs}\prod_{i<j}\mathcal{A}_{ij}^{rs}!\prod_i\mathcal{A}_{ii}^{rs}!!\prod_re_r!}\\
    P(\bm{k}|\bm{e}) &= \prod_r\multiset{e_r}{N}^{-1}\label{eq:kflat_prior}\\
  P(\bm{e}|\bar\omega) &= \prod_{r\le s}\frac{\bar\omega^{e_{rs}}}{(\bar\omega+1)^{e_{rs}+1}} = \frac{\bar\omega^E}{(\bar\omega+1)^{E+B(B+1)/2}}.
\end{align}
where $e_{rs}=\sum_{ij}\mathcal{A}_{ij}^{rs}$ is the total
number of edges between groups $r$ and $s$ (we used the shorthand
$e_r=\sum_se_{rs}$ and $k_i^r=\sum_{js}\mathcal{A}_{ij}^{rs}$), $P(\bm{\mathcal{A}}|\bm{k},\bm{e})$ is the probability of a
labeled graph $\bm{\mathcal{A}}$ where the labeled degrees $\bm{k}$
and edge counts between groups $\bm{e}$ are constrained to specific
values (and not their expectation values),  $P(\bm{k}|\bm{e})$ is the uniform prior distribution of the labeled
degrees constrained by the edge counts $\bm{e}$, and
$P(\bm{e}|\bar\omega)$ is the prior distribution of edge counts, given by a
mixture of independent geometric distributions with average
$\bar\omega$. 

The main advantage of this alternative model formulation is that it
allows us to remove the homogeneous assumptions by replacing the uniform
priors $P(\bm{k}|\bm{e})$ and $P(\bm{e}|\bar\omega)$ by a hierarchy of
priors and hyperpriors that incorporate the possibility of higher-order
structures. This can be achieved in a tractable manner without the need
of solving complicated integrals that would be required by introducing
deeper Bayesian hierarchies in Eq.~\eqref{eq.marginal} directly.

In a first step, we follow the approach of
Ref.~\cite{peixoto.2015} and condition the labeled degrees
$\bm{k}$ on an overlapping partition $\bm{b}=\{b_{ir}\}$, given by
\begin{equation}
  b_{ir} = \begin{cases}1 &\text{ if } k_i^r>0,\\0 &\text{ otherwise, }\end{cases}
\end{equation}
such that they are sampled by a distribution
\begin{equation}
  \label{eq.degrees}
  P(\bm{k}|\bm{e}) = P(\bm{k}|\bm{e},\bm{b}) P(\bm{b}).
\end{equation}
Importantly, the labeled degree sequence is sampled conditioned on the
\emph{frequency of degrees} $\bm{n}_{\bm{k}}^{\bm{b}}$ inside each
mixture $\bm{b}$, which itself is sampled from its own noninformative
prior,
\begin{equation}\label{eq:kprior}
  P(\bm{k}|\bm{e},\bm{b}) = \left[\prod_{\bm{b}}P(\bm{k}_{\bm{b}}|\bm{n}_{\bm{k}}^{\bm{b}})P(\bm{n}_{\bm{k}}^{\bm{b}}|\bm{e}_{\bm{b}},\bm{b})\right]P(\bm{e}_{\bm{b}}|\bm{e},\bm{b}),
\end{equation}
where $\bm{e}_{\bm{b}}$ is the number of incident edges in each mixture
(for detailed expressions see \textit{Supplementary Materials} Sec. 1.3).

Due to the fact that the frequencies of the mixtures as well as the
frequencies of the labeled degrees are treated as latent variables,
this model admits group mixtures which are far more heterogeneous than
the Dirichlet prior used in LDA. In particular, as was shown in
Ref.~\cite{peixoto_nonparametric_2017}, the expected degrees generated
in this manner follow a Bose-Einstein distribution, which is much
broader than the exponential distribution obtained with the prior of
Eq.~\eqref{eq:kflat_prior}. More importantly, the asymptotic form of the
degree likelihood will approach the true distribution as the prior
washes out~\cite{peixoto_nonparametric_2017}, making it more suitable
for skewed empirical frequencies, such as Zipf's law or mixtures
thereof~\cite{gerlach.2013}, without requiring specific parameters ---
such as exponents --- to be determined a priori.

In a second step, we follow
Refs.~\cite{peixoto.2014a,peixoto_nonparametric_2017} and
model the prior for the edge counts $\bm{e}$ between groups by
interpreting it as an adjacency matrix itself, i.e. a multigraph where
the $B$ groups are the nodes.  We then proceed by generating it from
another SBM which, in turn, has its own partition into groups and matrix of edge
counts.  Continuing in the same manner yields a hierarchy of nested
SBMs, where each level $l=1,\ldots,L$ clusters the
groups of the levels below.  This yields a probability (see
Ref.~\cite{peixoto_nonparametric_2017}) given by
\begin{equation}
  P(\bm{e}|E) = \prod_{l=1}^LP(\bm{e}_l|\bm{e}_{l+1},\bm{b}_l)P(\bm{b}_l)
\end{equation}
with
\begin{align}
  P(\bm{e}_l|\bm{e}_{l+1},\bm{b}_l) &= \prod_{r<s}\multiset{n_r^ln^l_s}{e_{rs}^{l+1}}^{-1}
  \prod_{r}\multiset{n_r^l(n_r^l+1)/2}{e_{rr}^{l+1}/2}^{-1} \\
  P(\bm{b}_l) &= \frac{\prod_rn_r^l!}{B_{l-1}!}{B_{l-1}-1\choose B_l-1}^{-1}\frac{1}{B_{l-1}}\label{eq:hpartition},
\end{align}
where the index $l$ refers to the variable of the SBM at a particular
level, e.g., $n_r^l$ is the number of nodes in group $r$ at level $l$.

The use of this hierarchical prior is a strong departure from the
noninformative assumption considered previously while containing it as a special case  when the depth of the hierarchy is $L=1$.  
It means
that we expect some form of heterogeneity in the data at multiple
scales, where groups of nodes are themselves grouped in larger groups
forming a hierarchy.  Crucially, this removes the ``unimodality''
inherent in the LDA assumption, as the group mixtures are now modeled
by another generative level which admits as much heterogeneity as the
original one. Furthermore, it can be shown to significantly alleviate
the resolution limit of the noninformative approach, since it enables
the detection of at most $O(N/\log N)$ groups in a sparse network with
$N$ nodes~\cite{peixoto.2014a,peixoto_nonparametric_2017}.

Given the above model we can find the best overlapping partitions of the
nodes by maximizing the posterior distribution
\begin{align}
  P(\{\bm{b}_l\}|\bm{A}) &= \frac{P(\bm{A},\{\bm{b}_l\})}{P(\bm{A})},
\end{align}
with
\begin{align}\label{eq:joint}
  P(\bm{A},\{\bm{b}_l\}) = P(\bm{\mathcal{A}}|\bm{k},\bm{e}_1,\bm{b}_0)P(\bm{k}|\bm{e}_1,\bm{b}_0)P(\bm{b}_0)\times
  \prod_lP(\bm{e}_l|\bm{e}_{l+1},\bm{b}_l)P(\bm{b}_l).
\end{align}
which can be efficiently inferred using Markov Chain Monte Carlo, as described in
Refs.~\cite{peixoto.2015,peixoto_nonparametric_2017}. 
The
nonparametric nature of the model makes it possible to infer i) the depth of the hierarchy (containing the ``flat'' model in case the data does not support a hierarchical structure) and ii) the number
of groups for both documents and words directly from the posterior
distribution, without the need for extrinsic methods or supervised
approaches to prevent overfitting. The latter can be seen interpreting Eq.~\eqref{eq:joint} as a description length, see
discussion after Eq.~\eqref{eq.sigma}.

The model above generates arbitrary multigraphs, whereas text is
represented as a bipartite network of words and documents. Since the
latter is a special case of the former, where words and documents belong
to distinct groups, the model can be used as it is, as it will ``learn''
the bipartite structure during inference. However, a more consistent
approach for text is to include this information in the prior, since we
should not have to infer what we already know. This can be done via a
simple modification of the model, where one replaces the prior for the
overlapping partition appearing in Eq.~\eqref{eq.degrees} by
\begin{equation}
  P(\bm{b}) = P_{\text{w}}(\bm{b}^{\text{w}})P_{\text{d}}(\bm{b}^{\text{d}}),
\end{equation}
where $P_{\text{w}}(\bm{b}^{\text{w}})$ and
$P_{\text{d}}(\bm{b}^{\text{d}})$ 
now correspond to a disjoint overlapping
partition of the words and documents, respectively. Likewise, the same
must be done at the upper levels of the hierarchy, by replacing Eq.~\eqref{eq:hpartition} with
\begin{equation}
  P(\bm{b}_l) = P_{\text{w}}(\bm{b}^{\text{w}}_l)P_{\text{d}}(\bm{b}^{\text{d}}_l).
\end{equation}
In this way, by construction, words and documents will never be placed
together in the same group.


  \subsection{Comparing LDA and hSBM in real and artificial data}
  In this section we show that the theoretical considerations discussed in the previous section are relevant in practice. We show that hSBM constitutes a better model than LDA in three classes of problems. First, we construct simple examples that show that LDA fails in cases of non-Dirichlet topic mixtures, while hSBM is able to infer, both, Dirichlet and non-Dirichlet mixtures. Second, we show that hSBM outperforms LDA even in artificial corpora drawn from the generative process of LDA. Third, we consider five different real corpora. We perform statistical model selection based on the principle of minimum description length~\cite{rissanen_modeling_1978} and computing the description length $\Sigma$ (the smaller the better) of each model (for details see \textit{Materials and Methods}, Minimum Description Length).

\subsubsection{Failure of LDA in the case of non-Dirichlet mixtures}
The choice of the Dirichlet distribution as a prior for the topic mixtures $\bm{\theta}_d$ implies that the ensemble of topic mixtures $P(\bm{\theta}_d)$ is assumed to be either unimodal or concentrated at the edges of the simplex. This is an undesired feature of this prior because there is no reason why data should show these characteristics. In order to explore how this affects the inference of LDA, we construct a set of simple examples with $K=3$ topics which allow for easy visualization. Besides real data, we consider synthetic data constructed from the generative process of LDA --- in which case $P(\bm{\theta}_d)$ indeed follows a Dirichlet distribution --- and from cases in which the Dirichlet assumption is violated --- e.g. by superimposing two Dirichlet mixtures resulting in a bimodal instead of a unimodal  $P(\bm{\theta}_d)$.

\begin{figure*}[bt]
\centerline{\includegraphics[width=2.\columnwidth]{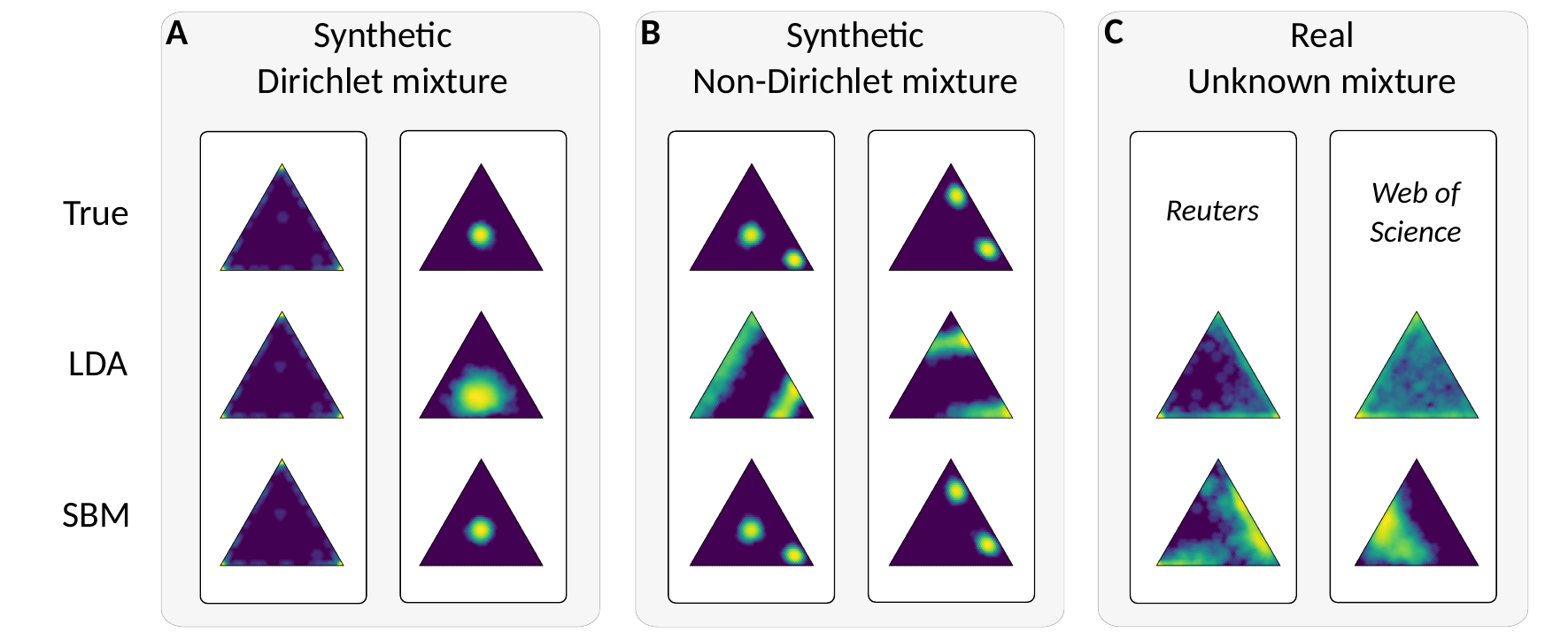}}
\caption{
LDA is unable to infer non-Dirichlet topic mixtures.
Visualization of the distribution of topic mixtures $\log P(\bm{\theta}_d)$  for different synthetic and real datasets in the 2-simplex using $K=3$ topics.
We show the true distribution in case of the synthetic data (top row) and the distributions inferred by LDA (middle row) and SBM (bottom row).
(A) Synthetic datasets with Dirichlet mixtures from the generative process of LDA with document hyperparameters $\bm{\alpha}_{d}=0.01\times(1/3,1/3,1/3)$ (left) and $\bm{\alpha}_{d}=100\times(1/3,1/3,1/3)$ (right) leading to different true mixture distributions $\log P(\bm{\theta}_d)$.  We fix the word hyperparameter $\beta_{rw}=0.01$, $D=1000$ documents, $V=100$ different words, and text length $k_d=1000$.
(B) Synthetic datasets with non-Dirichlet mixtures from combination of two Dirichlet mixture, respectively: $\bm{\alpha}_{d}\in \{ 100\times(1/3,1/3,1/3),100\times(0.1,0.8,0.1) \}$ (left) and $\bm{\alpha}_{d}\in \{ 100\times(0.1,0.2,0.7),100\times(0.1,0.7,0.2) \}$ (right).
(C) Real datasets with unknown topic mixtures: Reuters (left) and Web of Science (right) each containing $D=1000$ documents.
For LDA we use hyperparameter optimization.
For SBM we use an overlapping, nonnested parametrization in which each document belongs to its own group such that $B=D+K$ allowing for an unambiguous interpretation of the group membership as topic mixtures in the framework of topic models.
}
\label{fig.failure_mixtures}
\end{figure*}

The results summarized in Fig.~\ref{fig.failure_mixtures} show that SBM leads to better results than LDA. In Dirichlet generated data (Fig.~\ref{fig.failure_mixtures}A),  LDA self-consistently identifies the distribution of mixtures correctly.
Remarkably, the SBM is also able to correctly identify the Dirichlet mixture even though we did not explicitly specify Dirichlet priors. 
In the non-Dirichlet synthetic data  (Fig.~\ref{fig.failure_mixtures}B), the SBM results again closely match the true topic mixtures but LDA completely fails. In fact, although the inferred result by LDA no longer resembles a Dirichlet distribution after being influenced by data, it is significantly distorted by the unsuitable prior assumptions. 
Turning to real data (Fig.~\ref{fig.failure_mixtures}C), the LDA and SBM yield very different results. While the ``true'' underlying topic mixture of each document is unknown in this case, we can identify the negative consequence of the Dirichlet priors from the fact that the results from LDA are again similar to the ones expected from a Dirichlet distribution --- thus likely an artifact --- while the SBM results suggests a much richer pattern.

Taken together, the results of this simple example visually show that LDA not only struggles to infer non-Dirichlet mixtures, but also that it shows strong biases in the inference towards Dirichlet-type mixtures. On the other hand, SBM is able to capture a much richer spectrum of topic mixtures due to its nonparametric formulation. This is a direct consequence of the choice of priors: while LDA assumes \textit{a priori} that the ensemble of topic mixtures, $P(\bm{\theta}_d)$, follows a Dirichlet distribution, SBM is more agnostic with respect to the type of mixtures while retaining its nonparametric formulation.

%


\subsubsection{Artificial corpora sampled from LDA}
\label{sec.compare.artificial}

We consider artificial corpora constructed from the generative process
of LDA, incorporating some aspects of real texts, (for details see 
\textit{Materials and Methods}, Artificial corpora and \textit{Supplementary Materials} Sec. 2.1). Although LDA
is not a good model for real corpora --- as the Dirichlet
assumption is not realistic --- it serves to illustrate that even in a
situation that clearly favors LDA, the hSBM frequently provides a better
description of the data.

From the generative process we know the true latent variable of each
word-token.  Therefore, we are able to obtain the inferred topical
structure from each method by simply assigning the true labels without
using approximate numerical optimization methods for the inference.
This allows us to separate intrinsic properties of the model itself from external properties related to the numerical implementation.

In order to allow for a fair comparison between hSBM and LDA, we
consider two different choices in the inference of each method,
respectively. LDA requires the specification of a set of
hyperparameters $\bm{\alpha}$ and $\bm{\beta}$ used in the inference.
While in this particular case we know the \textit{true} hyperparameters
that generated the corpus, in general these are unknown.  Therefore, in
addition to the true values, we also consider a noninformative choice,
i.e. $\alpha_{dr}=1$ and $\beta_{rd}=1$. For the inference with hSBM,
we only use the special case where the hierarchy has a single level such that the prior is noninformative.  
We consider two different
parametrizations of the SBM: 1. Each document is assigned to its own
group, i.e. they are not clustered and 2. different documents can belong
to the same group, i.e. they are clustered.  While the former is
motivated by the original correspondence between pLSI and SBM, the
latter shows the additional advantage offered by the possibility of
clustering documents due to its symmetric treatment of words and
documents in a bipartite network (for details see \textit{Supplementary Materials} Sec. 2.2). 

%
\begin{figure}
\centerline{\includegraphics[width=1\columnwidth]{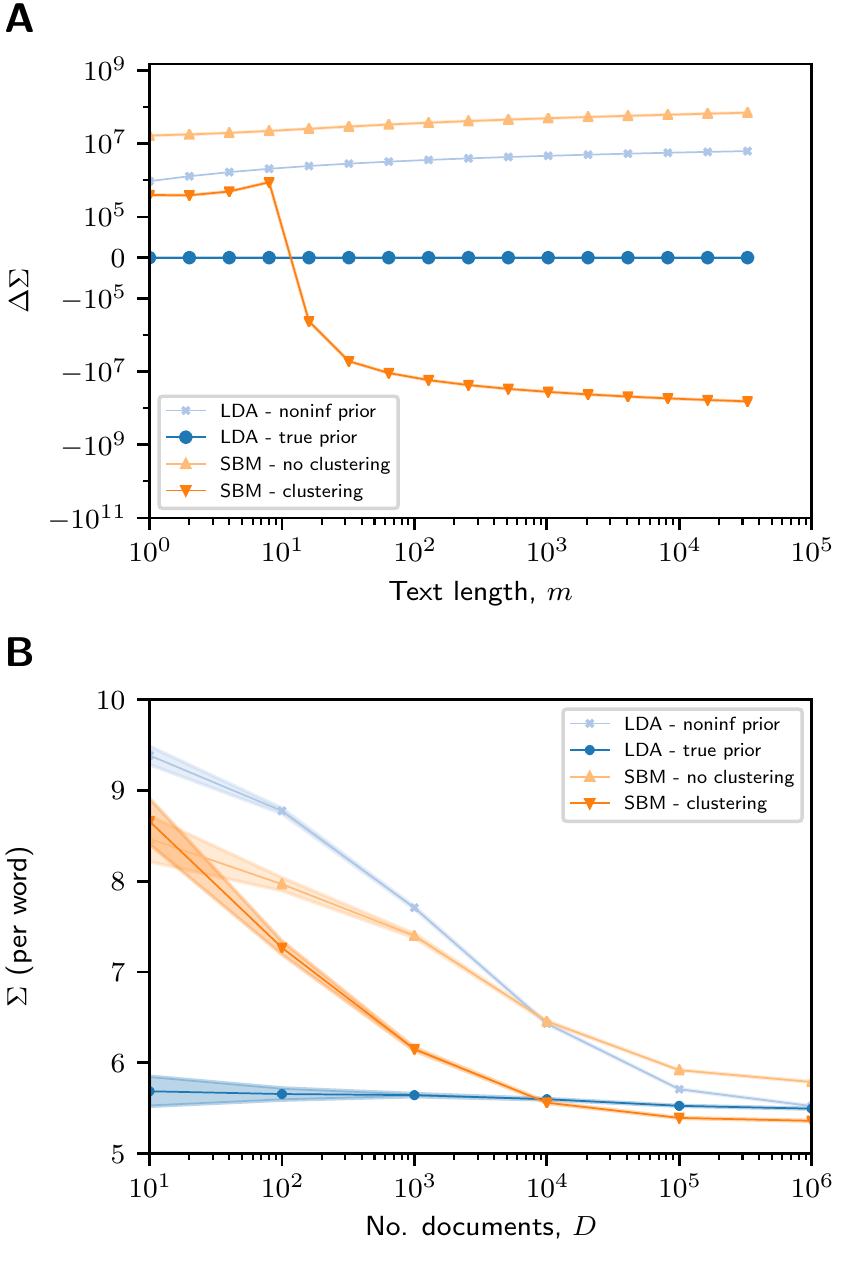}}
\caption{Comparison between LDA and SBM for artificial corpora drawn from LDA.
Description length $\Sigma$ of LDA and hSBM for an artificial corpus drawn from the generative process of LDA with $K=10$ topics.
(A) Difference in $\Sigma$, $\Delta \Sigma = \Sigma_i - \Sigma_{\text{LDA - true prior}}$, compared to the LDA with true priors --- the model that generated the data --- as a function of the text length $k_d=m$ and $D=10^6$ documents.
(B) Normalized $\Sigma$ (per word), as a function of the number of documents $D$ for fixed text length $k_d=m=128$.
The 4 curves correspond to different choices in the parametrization of the topic models:
i) LDA with noninformative priors (light blue - $\times$),
ii) LDA with true priors, i.e. the hyperparameters used to generate the artificial corpus (dark blue - $\bullet$), 
iii) hSBM with without clustering of documents (light orange - $\blacktriangle$),
and iv) hSBM with clustering of documents (dark orange - $\blacktriangledown$). 
\label{fig.plantedLDA}}
\end{figure}

In  Fig.~\ref{fig.plantedLDA}A, we show that hSBM is
consistently better than LDA for synthetic
corpora of almost any text length $k_d = m $ ranging over 4 orders of magnitude.
These results hold for asymptotically large corpora (in terms of the number of
documents) as shown in Fig.~\ref{fig.plantedLDA}B,
where we observe that the normalized description length of each model converges to
a fixed value when increasing the size of the corpus.
We confirm that these results hold across a wide range of parameter settings varying the number of topics as well as the values and base measures of the hyperparameters (\textit{Supplementary Materials} Sec. 3, Figs.~S1 - S3).

The LDA description length $\Sigma_{\text{LDA}}$ does not depend strongly on the considered prior (true or
noninformative) as the size of the corpora increases (Fig.~\ref{fig.plantedLDA}B).  This is
consistent with the typical expectation that in the limit of large data, the prior ``washes out''.  Note,
however, that for smaller corpora the $\Sigma$ of the
noninformative prior is significantly worse than the $\Sigma$ of the true
prior.

In contrast, the hSBM provides much shorter description lengths than LDA
for the same data when allowing documents to be clustered as well.  The
only exception is for very small texts ($m<10$ tokens) --- where we have
not converged to the asymptotic limit in the per-word description
length.  In the limit $D \rightarrow \infty$ we expect hSBM to provide a
similarly good or better model than LDA for all text lengths.  The
improvement of the hSBM over LDA in a LDA-generated corpus is
counterintuitive because, for sufficient data, we expect the true model
to provide a better description for it.  However, for a model like LDA
the limit of ``sufficient data'' involves the simultaneous scaling of
the number of documents, words, \emph{and} topics to very high values.
In particular, the generative process of LDA requires a large number of
documents to resolve the underlying Dirichlet distribution of
the topic-document distribution as well as a large number of topics to resolve the underlying word-topic distribution.  While the
former is realized growing the corpus by adding documents, the latter
aspect is nontrivial because the observed size of the vocabulary $V$ is
not a free parameter but is determined by the word-frequency
distribution and the size of the corpus through the so-called Heaps'
law~\cite{altmann.book2016}.  This means that as we grow the corpus by
adding more and more documents, initially the vocabulary increases linearly and only at very large corpora it 
settles into an asymptotic sublinear growth (\textit{Supplementary Materials} Sec. 4, Fig.~S4).  This, in
turn, requires an ever larger number of topics to resolve the underlying
word-topic distribution.  Such large number of topics is not feasible in
practice because it renders the whole goal and concept of topic models
obsolete --- compressing the information by obtaining an
effective, coarse-grained, description of the corpus at a manageable
number of topics.  

In summary, the limits in which LDA provides a better description, that is either extremely small
texts or very large number of topics,  are irrelevant in practice.
The observed limitations of
LDA are due to the following reasons: i) the finite number of topics
used to generate the data always leads to an under-sampling of the
Dirichlet distributions, and ii) LDA is redundant in the way it
describes the data in this sparse regime. In contrast, the assumptions
of the hSBM are better suited for this sparse regime, and hence leads to
a more compact description of the data, despite the fact the corpora
were in fact generated by LDA.


\subsubsection{Real corpora}
\label{sec.compare.real}
We compare LDA and SBM for a variety of different datasets, as shown in Table~\ref{tab.comparison} (for details see
\textit{Materials and Methods} Datasets for real corpora/Numerical implementations).
When using LDA, we consider both noninformative priors and fitted
hyperparameters, for a wide range of numbers of topics.  We obtain
systematically smaller values for the description length using the hSBM.
For real corpora, the difference is exacerbated by the fact the hSBM is
capable of clustering documents, capitalizing on a source of structure
in the data which is completely unavailable to LDA.

As our examples also show, LDA cannot be used in a direct manner to
choose the number of topics, as the noninformative choice systematically
underfits ($\Sigma_{\text{LDA}}$ increases monotonically with the number
of topics), and the parametric approach systematically overfits
($\Sigma_{\text{LDA}}$ decreases monotonically with the number of
topics). In practice, users are required to resort to
heuristics~\cite{arun_finding_2010,cao_density-based_2009}, or more
complicated inference approaches based on the computation of the model
evidence, which are not only numerically expensive, but can only be done
under onerous
approximations~\cite{griffiths.2004,wallach.2009a}.
In contrast, the hSBM is capable of extracting the appropriate number of
topics directly from its posterior distribution, while simultaneously
avoiding both under- and
overfitting~\cite{peixoto.2014a,peixoto_nonparametric_2017}.

In addition to these formal aspects, we argue that the hierarchical
nature of the hSBM, and the fact that it clusters words \emph{as well}
as documents, makes it more useful in interpreting text. We illustrate
this with a case study in the next section.

\begin{table*}
  \begin{minipage}{2.0\columnwidth}
\resizebox{1\textwidth}{!}{
  \begin{tabular}{l| rrr || rrrr || rrrr || r | r r}\hline
\multicolumn{4}{c||}{ \textbf{Corpus} } & \multicolumn{4}{c||}{$\Sigma_{\text{LDA}}$} & \multicolumn{4}{c||}{$\Sigma_{\text{LDA}}$ (hyperfit)}
& $\Sigma_{\text{hSBM}}$ & \multicolumn{2}{c}{hSBM groups}\\
\hline\hline
&Docs. & Words & Word Tokens & $10$ & $50$& $100$& $500$ &  $10$& $50$& $100$& $500$ & & Doc. & Words\\
\hline

Twitter         & 10,000 & 12,258 & 196,625   & 1,231,104  & 1,648,195 & 1,960,947 & 2,558,940 & 1,040,987 & 1,041,106 & 1,037,678 & 1,057,956 & \textbf{963,260} & 365 & 359 \\
Reuters         & 1,000  & 8,692  & 117,661   & 498,194    & 593,893 & 669,723 & 922,984 &   463,660 &   477,645 &   481,098 &   496,645 & \textbf{341,199} & 54 & 55 \\
Web of Science  & 1,000  & 11,198 & 126,313   & 530,519  & 666,447 & 760,114 & 1,056,554 & 531,893 & 555,727 & 560,455 & 571,291 & \textbf{426,529} & 16 & 18  \\
New York Times  & 1,000  & 32,415 & 335,749   & 1,658,815  & 1,673,333  & 2,178,439  & 2,977,931  & 1,658,815  & 1,673,333  & 1,686,495 & 1,725,057 & \textbf{1,448,631} & 124 & 125 \\
PlosONE         & 1,000  & 68,188 & 5,172,908 & 10,637,464 & 10,964,312 & 11,145,531 & 13,180,803 & 10,358,157 & 10,140,244 & 10,033,886 & 9,348,149 & \textbf{8,475,866} & 897 & 972
\end{tabular}}
\end{minipage}
\caption{Hierarchical SBM outperforms LDA in real corpora.
Each row corresponds to a different dataset (for details,  see Material \& Methods,
Datasets for real corpora).  We provide basic statistics of each dataset in columns
``Corpus''. The models are compared based on their description length $\Sigma$, see
Eq.~\eqref{eq.sigma}. We highlight the smallest $\Sigma$ for each corpus in bold in order to indicate the best model.
Results for LDA with noninformative and fitted hyperparameters are
shown in columns ``$\Sigma_{\text{LDA}}$'' and ``$\Sigma_{\text{LDA}}$ (hyperfit)'' for
different  number of topics $K \in \{ 10,50,100,500\}$. Result for the hSBM are shown in
column ``$\Sigma_{\text{hSBM}}$'' and the inferred number of groups (documents and words)
in ``hSBM groups''.\label{tab.comparison}\vspace{1em}}

\end{table*}


\subsection{Case study: Application of hSBM to Wikipedia articles}
\label{sec.compare.wiki}
\begin{figure*}[bt]
\centerline{\includegraphics[width=2\columnwidth]{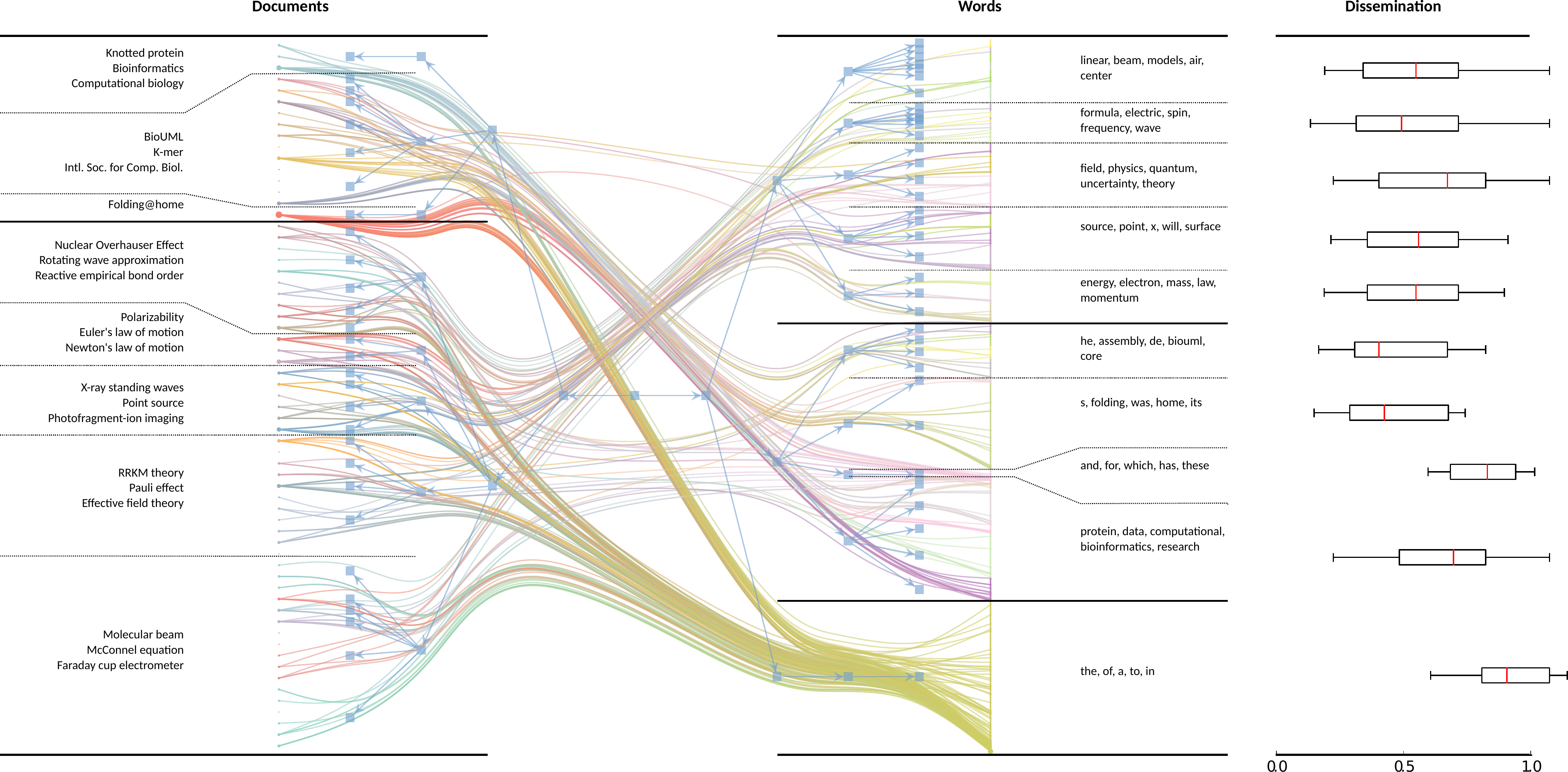}}
\caption{
Inference of hSBM  to articles from the Wikipedia. 
Articles from 3 categories (Chemical Physics, Experimental Physics, and Computational Biology).
The first hierarchical level reflects bipartite nature of the network with document-nodes (left) and word-nodes (right).
The grouping on the second hierarchical level is indicated by solid lines.
We show examples for nodes that belong to each groups on the third hierarchical level (indicated by dotted lines):
For word-nodes, we show the 5 most frequent words; for document-nodes, we show 3 (or fewer) randomly selected articles.
For each word, we calculate the dissemination coefficient $U_D$ which quantifies how unevenly words are distributed among documents~\cite{altmann.2011}: $U_D=1$ indicates the expected dissemination from a random null model;  the smaller $U_D$ ($0<U_D<1$), the more unevenly a word is distributed.  
We show the $5,25,50,75,95$-percentile for each group of word-nodes on the third level of the hierarchy.
}
\label{fig.wikipedia}
\end{figure*}

We illustrate the results of the inference with the hSBM for articles
taken from the English Wikipedia in Fig.~\ref{fig.wikipedia}, showing
the hierarchical clustering of documents and words. To make the
visualization clearer, we focus on a small network created from only
three scientific disciplines: Chemical Physics (21 articles),
Experimental Physics (24 articles), and Computational Biology (18
articles).  For clarity, we only consider words that appear more than
once, such that we end up with a network of $63$ document-nodes, $3,140$
word-nodes, and $39,704$ edges.

The hSBM splits the network into groups on different levels, organized
as a hierarchical tree. Note that the number of groups and the number of
levels were not specified beforehand but automatically detected in the
inference.  On the highest level, hSBM reflects the bipartite structure
into word- and document-nodes, as is imposed in our model.  

  In contrast to traditional topic models such as LDA, hSBM automatically clusters documents into groups.
While we considered articles from three different categories (one category from
biology and two categories from physics), the second level in the
hierarchy separates documents into only two groups corresponding to
articles about biology (e.g. bioinformatics or K-mer) and articles on
physics (e.g. Rotating wave approximation or Molecular beam).  For lower
levels, articles become separated into a larger number of groups,
e.g. one group contains two articles on Euler's and Newton's law of
motion, respectively. 

For words, the second level in the hierarchy splits nodes into three separate groups.  We find that two groups
represent words belonging to physics (e.g. beam, formula, or energy) and biology (assembly, folding, or protein) while the third group represents function words (the, of, or a). 
In fact, we find that the latter group's words show close-to random distribution across documents by calculating the dissemination coefficient (right side of Fig.~\ref{fig.wikipedia}, see caption for definition).
Furthermore, the median dissemination of the other groups  is substantially less random with the exception of one subgroup (containing and, for, or which).
This suggests a more data-driven approach to dealing with function words in topic models.
The standard practice is to remove words from a manually curated list of stopwords, however,  recent results question the efficacy of such methods~\cite{Schoffield2017}.
In contrast, the hSBM is able to automatically identify groups of stopwords, potentially rendering such heuristic interventions unnecessary.

\section{Discussion}
\label{sec.discuss}

The underlying equivalence between pLSI and the overlapping version of
the SBM means that the ``bag of words'' formulation of topical corpora
is mathematically equivalent to bipartite networks of words and
documents with modular structures.  From this we were able to formulate
a topic model based on a hierarchical version of the SBM (hSBM) in a
fully Bayesian framework alleviating some of the most serious conceptual
deficiencies in current approaches to topic modeling such as LDA.  In
particular, the model formulation is nonparametric, and model complexity
aspects such as the number of topics can be inferred directly from the
model's posterior distribution. Furthermore, the model is based on a
hierarchical clustering of both words and documents --- in contrast to
LDA which is based on a nonhierarchical clustering of the words alone.
This enables the identification of structural patterns in text that is
unavailable to LDA, while at the same time allowing for the
identification of patterns in multiple scales of resolution.

We have shown that hSBM constitutes a better topic model compared to LDA
not only for a diverse set of real corpora but even for artificial
corpora generated from LDA itself.  
It is capable of providing better compression  --- as a measure of the quality of fit --- as well as a richer interpretation of the data. 
More importantly, however, the hSBM offers an alternative to Dirichlet priors employed in virtually any variation of current
approaches to topic modeling.  While motivated by their computational
convenience, Dirichlet priors do not reflect prior knowledge compatible with the
actual usage of language.
In fact, our analysis suggests that Dirichlet priors introduce severe biases into the inference result, which in turn dramatically hinder its performance in case of even just slight deviations from the Dirichlet assumption.
In contrast, our work shows how to formulate and incorporate different (and as we have shown more suitable) priors in a fully Bayesian framework, which are completely agnostic to the type of inferred mixtures.
Furthermore, it also serves as a working example that efficient numerical
implementations of non-Dirichlet topic models are feasible and can be
applied in practice to large collections of documents.

More generally, our results show how the same mathematical ideas can be
used to two extremely popular and mostly disconnected problems: the
inference of topics in corpora and of communities in networks. We used
this connection to obtain improved topic models, but there are many
additional theoretical results in community detection that should be
explored in the topic model context, e.g., fundamental limits to
inference such as the undetectable-detectable phase
transition~\cite{decelle.2011} or the analogy to Potts-like spin systems
in statistical physics~\cite{hu.2012}.  Furthermore, this
connection allows the many extensions of the SBM, such as
multilayer~\cite{peixoto_inferring_2015} and annotated~\cite{newman_structure_2016,hric_network_2016} versions to be readily used for
topic modeling of richer text including hyperlinks, citations between
documents, etc. 
Conversely, 
the field of topic modeling has long adopted a Bayesian perspective to inference, which until now has not seen a widespread use in community detection.
Thus, insights from topic modeling about either the formulation of suitable priors, or the approximatinon of posterior distributions, might catalyze the development of improved statistical methods to detect communities in networks.
Furthermore, the traditional application of topic models in the analysis of texts leads to classes of networks usually not considered by
community detection algorithms. The word-document network is bipartite (words-documents), the topics/communities can be overlapping, and the
number of links (word-tokens) and nodes (word-types) are connected to
each other through Heaps' law.  
In particular, the latter aspect results in dense networks, which have been largely overlooked by the networks community~\cite{Courtney2018}.
Topic models, thus, might provide additional insights how to approach such networks as it remains unclear how such properties affect the inference of communities in word-document networks.
More generally, Heaps' law constitutes only one of numerous statistical laws in language~\cite{altmann.book2016}, such as the well-known Zipf's law~\cite{zipf.1936}. While these regularities are well-studied empirically, few attempts have been made to incorporate them explicitly as prior knowledge, e.g. formulating generative processes that lead to Zipf's law~\cite{Sato2010,Buntine2014}.
Our results show that the SBM provides a flexible approach to deal with Zipf's law which constitutes a challenge to state-of-the-art topic models such as LDA.
Zipf's law appears also in genetic codes~\cite{Mantegna1994} and images~\cite{Sudderth2009}, two prominent fields in which LDA-type models have been extensively applied~\cite{pritchard.2000,Broderick2015}, suggesting that the block-model approach we introduce here is promising also beyond text analysis.

\section{Materials \& Methods}

\subsection{Minimum Description Length }
We compare both models based on the description length $\Sigma$, where smaller values indicate a better model~\cite{rissanen_modeling_1978}. 
We obtain $\Sigma$ for LDA from Eq.~\eqref{eq.marginalL} and $\Sigma$ for hSBM from
Eq.~\eqref{eq:joint} as 
\begin{align}\label{eq.sigma}
  \Sigma_{\text{LDA}} &= -\ln P(\bm{n}|\bm{\eta}, \bm{\beta}, \bm{\alpha})P(\bm{\eta})\\
  \Sigma_{\text{hSBM}} &= -\ln P(\bm{A},\{\bm{b}_l\}).
\end{align}
We note that $\Sigma_{\text{LDA}}$  is conditioned on the hyperparameters $\bm{\beta}$, $\bm{\alpha}$ and therefore it is exact
for noninformative priors ($\alpha_{dr}=1$ and $\beta_{rd}=1$) only. Otherwise,
Eq.~\eqref{eq.sigma} is only a lower bound for $\Sigma_{\text{LDA}}$ because it lacks the
terms involving  \emph{hyperpriors} for $\bm{\beta}$ and $\bm{\alpha}$.
For simplicity, we  ignore this correction in our analysis and therefore we favor LDA. 

The motivation for this approach is two-fold.

One the one hand it offers a well-founded approach to unsupervised model selection within
the framework of information theory, as it corresponds to the amount of
information necessary to describe simultaneously i) the data when the
model parameters are known, and ii) the parameters themselves.  As the
complexity of the model increases, the former will typically decrease,
as it fits more closely the data, while at the same time it is
compensated by an increase of the latter term, which serves as a penalty
that prevents overfitting.  
In addition, given data and two models $M_1$ and $M_2$ with description length $\Sigma_{M_1}$ and
$\Sigma_{M_2}$, we can relate the difference $\Delta \Sigma \equiv
\Sigma_{M_1} - \Sigma_{M_2}$ to the Bayes' Factor (BF)~\cite{kass.1995}.
The latter quantifies how much more likely one model is compared to the other given the
data
\begin{equation}
\text{BF}  \equiv \frac{ P( M_1 \mid \text{data} )}{P( M_2 \mid \text{data})} = \frac{ P(\text{data}\mid M_1)P(M_1)}{P(\text{data}\mid M_2)P(M_2)} = e^{-\Delta \Sigma},
\end{equation}
where we assume that each model is a priori equally likely, i.e. $P(M_1)=P(M_2)$. 

The description length allows for a straightforward model
comparison without the introduction of confounding factors.  In fact,
commonly used supervised model selection approaches such as perplexity
require additional approximation techniques~\cite{wallach.2009a}, which
are not readily applicable to the microcanonical formulation of the SBM.
It is thus not clear whether any difference in predictive power would
result from the model and its inference or the approximation used in the
calculation of perplexity. Furthermore, we note that it has been shown
recently that supervised approaches based on the held-out likelihood of
missing edges tend to overfit in key cases, failing to select the most
parsimonious model, unlike unsupervised approaches which are more
robust~\cite{Valles-Catala2017}.

\subsection{Artificial corpora}
For the construction of the artificial corpora, we fix the parameters in
the generative process of LDA, i.e. the number of topics $K$, the
hyperparameters $\bm{\alpha}$ and $\bm{\beta}$, and the length of
individual articles $m$. 
The $\bm{\alpha}$ ($\bm{\beta}$) -
hyperparameters determine the distribution of topics (words) in each
document (topic).  

The generative process of LDA can be described in the following way.
For each topic $r\in\{1,\ldots,K\}$ we sample a
distribution over words $\bm{\phi}_r$ from a $V$-dimensional Dirichlet
distribution with parameters $\beta_{rw}$ for $w\in\{1,\ldots,V\}$.  
For each document $d\in\{1,\ldots,D\}$ we sample a topic mixture
$\bm{\theta}_{d}$ from a $K$-dimensional Dirichlet distribution with
parameters $\alpha_{dr}$ for $r\in\{1,\ldots,K\}$. For each word position 
$l_d \in\{1,\ldots,k_d\}$ ($k_d$ is the length of document
$d$) we first sample a topic $r^*=r_{l_d}$ from a multinomial with
parameters $\bm{\theta}_{d}$ and then sample a word $w$ from a multinomial
with parameters $\bm{\phi}_{r^*}$.

We assume a parametrization in which i) each document has the same
topic-document hyperparameter, i.e. $\alpha_{dr}=\alpha_r$ for $d\in\{1,\ldots,D\}$,
and ii) each topic has the same word-topic hyperparameter, i.e. $\beta_{rw}=\beta_w$ for $r\in\{1,\ldots,K\}$.  
We fix the average probability of occurrence of a topic, $p_r$, (word, $p_w$) by a
introducing scalar hyperparameters $\alpha$ ($\beta$), i.e. $\alpha_{dr}
= \alpha K (p_r)$ for $r\in\{1,\ldots,K\}$ ($\beta_{rw} = \beta V (p_w)$
for $w=1,\ldots,V$).  In our case we choose 
i) equiprobable topics, i.e. $p_r = 1/K$ and 
ii) empirically measured word frequencies from the Wikipedia corpus, i.e. $p_w =
p_w^{\mathrm{emp}}$ with $w=1, \ldots, 95129$, yielding a Zipfian distribution (\textit{Supplementary Materials} Sec. 5, Fig.~S5), shown to be universally
described by a double power law~\cite{gerlach.2013}.

\subsection{Datasets for real corpora}
For the comparison of hSBM and LDA we consider different datasets of
written texts varying in genre, time of origin, average text length,
number of documents, and language; as well as datasets used in previous
works on topic models,
e.g.~\cite{blei.2003,wallach.2009,asuncion.2009,lancichinetti.2015}:
\begin{enumerate}
\item ``Twitter'', a sample of Twitter messages obtained from \url{http://www.nltk.org/nltk_data/};
\item ``Reuters'', a collection of documents from the Reuters financial newswire service denoted as ``Reuters-21578, Distribution 1.0'' obtained from \url{http://www.nltk.org/nltk_data/};
\item ``Web of Science'', abstracts from physics papers published in the year 2000;
\item ``New York Times (NYT)'', a collection of newspaper articles obtained from \url{http://archive.ics.uci.edu/ml};
\item ``PlosOne'', full text of all scientific articles published in 2011 in the journal PLoS One obtained via the Plos API (\url{http://api.plos.org/})
\end{enumerate}
In all cases we considered a random subset of the documents, as detailed
in Table~\ref{tab.comparison}. For the NYT data we did not employ any
additional filtering since the data was already provided in the form of
pre-filtered word counts.  For the other datasets we employed the
following filtering: i) we decapitalized all words, ii) we replaced
punctuation and special characters (e.g. ``.'', ``,'', or ``/'') by
blank spaces such that we can define a word as any substring between two
blank spaces, and iii) keep only those words which consisted of the
letters a-z.

\subsection{Numerical Implementations}
For inference with LDA we used package mallet (\url{http://mallet.cs.umass.edu/}).  
The algorithm for inference with the hSBM presented in this work is
implemented in C++ as part of the \texttt{graph-tool} Python
library (\url{https://graph-tool.skewed.de}). 
We provide code on how to use hSBM for topic modeling in a github repository (\url{https://topsbm.github.io/}).

\begin{acknowledgments}
We thank M. Palzenberger for help with the ``Web of Science" data.
EGA thanks L. Azizi and W. L. Buntine for helpful discussions.

\paragraph*{Author contributions:}
M.G., T.P.P., and E.G.A designed research; 
M.G., T.P.P., and E.G.A performed research; 
M.G. and T.P.P. analyzed data; 
M.G., T.P.P., and E.G.A wrote the paper.
\paragraph*{Competing interests:}
The authors declare no competing interests.
\paragraph*{Data and materials availability:}
Data and code are available from the sources indicated above or from authors upon request.
\end{acknowledgments}

\bibliographystyle{ScienceAdvances}

\newpage
\clearpage
\begin{widetext}

\begin{center}
{\Large {\bf Supplementary Material for the manuscript:\\ ``A network approach to topic models'' \vspace{15pt}}} \\ 

\end{center}

\renewcommand{\theequation}{S\arabic{equation}}
\renewcommand{\thefigure}{S\arabic{figure}}
\renewcommand{\thetable}{S\arabic{table}}

\setcounter{equation}{0}  
\setcounter{figure}{0}
\setcounter{table}{0}
\setcounter{section}{0}

\maketitle

\section{Marginal likelihood of the SBM}

\subsection{Noninformative priors}

For the labeled network $\bm{\mathcal{A}}$ considered in the main text, section {\bf Community detection: The hierarchical SBM}, Eq.~(4), we have
\begin{align}
  P(\bm{\mathcal{A}}|\bm{\kappa},\bm{\omega}) &=
  \prod_{i<j}\prod_{rs}\frac{e^{-\kappa_{ir}\omega_{rs}\kappa_{is}}(\kappa_{ir}\omega_{rs}\kappa_{js})^{\mathcal{A}_{ij}^{rs}}}{\mathcal{A}_{ij}^{rs}!}\times\nonumber\\
  &\qquad\prod_i\prod_{rs}\frac{e^{-\kappa_{ir}\omega_{rs}\kappa_{is}/2}(\kappa_{is}\omega_{rs}\kappa_{is}/2)^{\mathcal{A}_{ii}^{rs}/2}}{\mathcal{A}_{ii}^{rs}/2!}.
\end{align}
If we now make a noninformative choice for the priors,
\begin{align}
  P(\bm{\kappa}) &= \prod_r(n-1)!\delta(\textstyle\sum_i\kappa_{ir}-1),\\
  P(\bm{\omega}|\bar{\omega}) &= \prod_{r\le s} \frac{e^{-\omega_{rs}/\bar{\omega}}}{\bar{\omega}},
\end{align}
we can compute the integrated marginal likelihood as
\begin{align}\label{eq.marginal.sm}
  P(\bm{\mathcal{A}}|\bar\omega) &= \int P(\bm{\mathcal{A}}|\bm{\kappa},\bm{\omega})P(\bm{\kappa})P(\bm{\omega}|\bar{\omega})\; \dd\bm{\kappa}\dd\bm{\omega},\nonumber\\
  &= \frac{\bar\omega^E}{(\bar\omega+1)^{E+B(B+1)/2}}\frac{\prod_{r<s}e_{rs}!\prod_re_{rr}!!}{\prod_{rs}\prod_{i<j}\mathcal{A}_{ij}^{rs}!\prod_i\mathcal{A}_{ii}^{rs}!!}\times\nonumber\\
  &\qquad \prod_r\frac{(N-1)!}{(e_r+N-1)!}\prod_{ir}k_i^r!.
\end{align}

\subsection{Equivalence with microcanonical model}
As mentioned in the main text, Eq.~(7) can be decomposed
as
\begin{equation}
  P(\bm{\mathcal{A}}|\bar\omega) =  P(\bm{\mathcal{A}},\bm{k},\bm{e}|\bar\omega) = P(\bm{\mathcal{A}}|\bm{k},\bm{e})P(\bm{k}|\bm{e})P(\bm{e}|\bar\omega),
\end{equation}
with
\begin{align}
  P(\bm{\mathcal{A}}|\bm{k},\bm{e}) &= \frac{\prod_{r<s}e_{rs}!\prod_re_{rr}!!\prod_{ir}k_i^r!}{\prod_{rs}\prod_{i<j}\mathcal{A}_{ij}^{rs}!\prod_i\mathcal{A}_{ii}^{rs}!!\prod_re_r!}\\
  P(\bm{k}|\bm{e}) &= \prod_r\multiset{e_r}{N}^{-1}\label{eq:kflat_prior.sm}\\
  P(\bm{e}|\bar\omega) &= \prod_{r\le s}\frac{\bar\omega^{e_{rs}}}{(\bar\omega+1)^{e_{rs}+1}} = \frac{\bar\omega^E}{(\bar\omega+1)^{E+B(B+1)/2}}.
\end{align}
where $e_{rs}=\sum_{ij}\mathcal{A}_{ij}^{rs}$ is the total
number of edges between groups $r$ and $s$ (we used the shorthand
$e_r=\sum_se_{rs}$ and $k_i^r=\sum_{js}\mathcal{A}_{ij}^{rs}$).
$P(\bm{\mathcal{A}}|\bm{k},\bm{e})$ is the probability of a
labelled graph $\bm{\mathcal{A}}$ where the labelled degrees $\bm{k}$
and edge counts between groups $\bm{e}$ are constrained to specific
values. This can be seen by writing
\begin{equation}
  P(\bm{\mathcal{A}}|\bm{k},\bm{e}) = \frac{\Xi}{\Omega},
\end{equation}
with
\begin{equation}
  \Omega = \frac{\prod_re_r!}{\prod_{r<s}e_{rs}!\prod_re_{rr}!!}
\end{equation}
being the number of configurations (i.e. half-edge pairings) that are
compatible with the constraints, and
\begin{equation}
  \Xi = \frac{\prod_{ir}k_i^r!}{\prod_{rs}\prod_{i<j}\mathcal{A}_{ij}^{rs}!\prod_i\mathcal{A}_{ii}^{rs}!!}
\end{equation}
is the number of configurations that correspond to the same labelled
graph $\{A_{ij}^{rs}\}$.  $P(\bm{k}|\bm{e})$ is the uniform prior
distribution of the labelled degrees constrained by the edge counts
$\bm{e}$, since $\multiset{e_r}{N}$ is the number of ways to distribute
$e_r$ indistinguishable items into $N$ distinguishable
bins. Furthermore, $P(\bm{e}|\bar\omega)$ is the prior distribution of
edge counts, given by a mixture of independent geometric distributions
with average $\bar\omega$.

\subsection{Labelled degrees and overlapping partitions}
As described in the main text, section {\bf Community detection: The hierarchical SBM}, Eq.~(13), the distribution of labeled degrees is
given by
\begin{equation}
  P(\bm{k}|\bm{e}) = P(\bm{k}|\bm{e},\bm{b}) P(\bm{b}),
\end{equation}
where the overlapping partition is distributed according to
\begin{equation}
  P(\bm{b}) = \left[\prod_qP(\bm{b}_q|\bm{n}_{\bm{b}}^q)P(\bm{n}_{\bm{b}}^q|n_q)\right]
  P(\bm{q}|\bm{n})P(\bm{n}).
\end{equation}
Here, $\bm{b}$ corresponds to a specific set of groups, i.e. a mixture,
of size $q=|\bm{b}|$. The distribution above means that we first sample
the frequency of mixture sizes from the distribution
\begin{equation}
  P(\bm{n}) = \multiset{Q}{N}^{-1},
\end{equation}
where $Q$ is the maximum overlap size (typically $Q=B$, unless we want
to force nonoverlapping partitions with $Q=1$). Given the frequencies,
the mixture sizes are sampled uniformly on each node
\begin{equation}
  P(\bm{q}|\bm{n}) = \frac{\prod_qn_q!}{N!}.
\end{equation}
We now consider the nodes with a given value of $q_i=q$ separately, and
we put each one of them in a specific mixture $\bm{b}$ of size $q$. We
do so by first sampling the frequencies in each mixture $\bm{n}_{\bm b}^q$
uniformly
\begin{equation}
P(\bm{n}_{\bm{b}}^q|n_q) = \multiset{{B\choose q}}{n_q}^{-1},
\end{equation}
and then we sample the mixtures themselves, conditioned on the
frequencies,
\begin{equation}
  P(\bm{b}_q|\bm{n}_{\bm{b}}^q) = \frac{\prod_{\bm{b}}n_{\bm b}^q!}{n_q!}.
\end{equation}
The labeled degree sequence is sampled conditioned on this overlapping
partition and also on the frequency of degrees
$\bm{n}_{\bm{k}}^{\bm{b}}$ inside each mixture $\bm{b}$,
\begin{equation}
  P(\bm{k}|\bm{e},\bm{b}) = \left[\prod_{\bm{b}}P(\bm{k}_{\bm{b}}|\bm{n}_{\bm{k}}^{\bm{b}})P(\bm{n}_{\bm{k}}^{\bm{b}}|\bm{e}_{\bm{b}},\bm{b})\right]P(\bm{e}_{\bm{b}}|\bm{e},\bm{b}).
\end{equation}
Here, $e^r_{\bm{b}}=\sum_ik_i^r\delta_{b_i^r,1}$ is the sum of the
degrees with label $r$ in mixture $\bm{b}$, which is sampled uniformly
according to
\begin{equation}
  P(\bm{e}_{\bm{b}}|\bm{e},\bm{b}) = \prod_r\multiset{m_r}{e_r}^{-1},
\end{equation}
where $m_r=\sum_{\bm{b}}b_r[n_{\bm{b}}>0]$ is the number of occupied
mixtures that contain component $r$. Given the degree sums, the
frequency of degrees is sampled according to
\begin{equation}
  P(\bm{n}_{\bm{k}}^{\bm{b}}|\bm{e}_{\bm{b}},\bm{b}) = \prod_{r\in\bm{b}}p(e_{\bm{b}}^r, n_{\bm{b}}^r)^{-1},
\end{equation}
where $p(m,n)$ is the number of partitions of the integer $m$ into
exactly $n$ parts, which can be pre-computed via the recurrence
\begin{equation}
  p(m, n) = p(m-n,n)+p(m-1,n-1),
\end{equation}
with the boundary conditions $p(0,0)=1$ and $p(m,n)=0$ if $n\le 0$ or
$m\le 0$,  or alternatively via the relation
\begin{equation}
  p(m+n, n) = q(m, n)
\end{equation}
where $q(m, n)$ is the number of partitions of $m$ into \emph{at most}
$n$ parts, and using accurate asymptotic approximations for $q(m,n)$
(see Ref.~\cite{peixoto_nonparametric_2017}). Finally, having sampled
the frequencies, we sample the labeled degree sequence uniformly in
each mixture
\begin{equation}
  P(\bm{k}_{\bm{b}}|\bm{n}_{\bm{k}}^{\bm{b}}) = \frac{\prod_{\bm{k}}n_{\bm{k}}^{\bm{b}}!}{n_{\bm{b}}!}.
\end{equation}
We refer to Ref.~\cite{peixoto.2015} for further details of the above distribution. 

\section{Artificial corpora drawn from LDA}
\subsection{Drawing artificial documents from LDA}
\label{sec.drawLDA}
We specify $\alpha_{dr}$ and $\beta_{rw}$, i.e. the hyperparameters used to \textit{generate} the artificial corpus (note that the hyperparameters used in the inference with LDA can be different)  and fixing $V$, $K$, $D$, $M$ and proceed in the following way:
\begin{itemize}
 \item For each topic $r=1,\ldots,K$:
 \begin{itemize}
 \item
 Draw the word-topic distribution $\phi_w^r$ (frequencies of words conditioned on the topic $r$) from a $V$-dimensional Dirichlet:\\
 $\phi_w^r \sim Dir_{V}(\beta_{wr})$
 \end{itemize}
 \item For each document $d=1,\ldots,D$:
  \begin{itemize}
  \item
  Draw the topic-document distribution $\theta_d^r$ (frequencies of topics conditioned on the doc $d$) from a $K$-dimensional Dirichlet:\\
 $\theta_d^r \sim Dir_{K}(\alpha_{dr})$
 \item For each token $i_d=1,\ldots,n_d$ ($n_d$ is the length of each document) in document $d$:
 \begin{itemize}
 \item Draw a topic $r_{i_d}$ from the categorical $\theta_d^r$
 \item Draw a word-type $w_{i_d}$ from the categorical $\phi_w^{r_{i_d}}$
 \end{itemize}
 \end{itemize}
\end{itemize}

\subsection{Inference of corpora drawn from LDA}
When we draw artificial corpora we obtain the labeled word-document counts $n_{wd}^r$, i.e. the ``true'' labels from the generative process of LDA as described above.
In the following we describe how to obtain the description length of LDA and SBM when assigning the ``true'' labels as the result of the inference.
In this way, we obtain the best possible inference results from each method.
We can, therefore, compare the two models conceptually and avoid the issue of which particular numerical implementation was used.

\subsubsection{Inference with LDA}
In the inference with LDA we simply need the word-topic, $n_w^r =
\sum_{d=1}^D n_{dw}^r$, the document-topic counts, $n_d^r=\sum_{w=1}^V
n_{dw}^r$, and the word-document matrix $n_{dw}=\sum_{r=1}^K n_{dw}^r$
and use them to obtain the description length for LDA.

Note that for the inference we also have to specify the hyperparameters used in the inference, $\hat{\alpha}_{dr}$ and $\hat{\beta}_{rw}$. 
One approach is to consider the \textit{true prior} (the same hyperparameter we used to generate the corpus) such that $\hat{\alpha}_{dr} = \alpha_{dr}$ and $\hat{\beta}_{rw}=\beta_{rw}$.
In general, however, the data is not generated from LDA such that it is unclear which is the best choice of hyperparameters for inference. Therefore, we also consider the case of a \textit{noninformative prior} in which $\hat{\alpha}_{dr} = 1$ and $\hat{\beta}_{rw}=1$. 

\subsubsection{Inference with SBM}
For the stochastic block model (SBM) we consider texts as a network in which the nodes consist of documents and words and the strength of the edge between them is given by the number of occurrences of the word in the document, yielding a bipartite multigraph.
We consider the case of a degree-corrected, overlapping SBM with only one layer in the hierarchy.

\paragraph{No clustering of documents}
For the SBM we use a particular parametrization starting from the equivalence between the degree-corrected SBM \cite{ball.2011} and probabilistic semantic indexing (pLSI) \cite{hofmann.1999}, as described in the main text, section {\bf Topic models: pLSI and LDA}
Each document-node is put in its own group and the word-nodes are clustered into word-groups.
The latter correspond to the topics in LDA (with possible mixtures among those groups) thus giving us a total of $B=D+K$ groups.

\paragraph{Clustering of documents}
Instead of putting each document in a separate group we cluster the documents into $K$ groups as well such that we have $B=2K$ groups in total. 
Note that this corresponds to a completely symmetric clustering of the groups in which we choose the indices such that $r=1,\ldots,K$ are groups for the document-nodes and $r=K+1,\ldots,2K$ are word-nodes.
For a given word-token of word-type $w$ appearing in document $d$ labeled in topic $r=j$, we label the two half-edges as $r_d = j$ (the half-edge on the document-node) and $r_w = K+j$ (the half-edge on the word-node).

\section{Varying the hyperparameters and number of topics }
In Fig.~4 of the main text we compare LDA and hSBM for corpora drawn from LDA for the case $K=10$ and $\alpha=\beta=1.0$. 
In Figs.~(\ref{fig.drawLDA.alphabeta}, \ref{fig.drawLDA.alphaK}, \ref{fig.drawLDA.as}) we show that these results hold under very general conditions by varying i) the values of the scalar hyperparameters; ii) the number of topics; and iii) the base measure of the vector-valued hyperparameters $\vec{\alpha}$ and $\vec{\beta}$ (symmetric or asymmetric following the approach in Ref.~\cite{wallach.2009}).
While the individual curves for the description length of the different models look different, the qualitative behavior shown in Fig.~4 of the main text remains the same.
In all cases, the hSBM performs better than the LDA with noninformative
priors; and only in few cases the hSBM has a larger description length
than LDA with the true hyperparameters which actually generated the
data.  Note that the latter case constitutes an exception because i) the
generating hyperparameters are unknown in practice; and ii) as the
hyperparameters deviate from the noninformative choice, the LDA
description length computed ceases to be complete, becoming only a lower
bound to the complete one which involves integration over the
hyperparameters (as is thus intractable).

\begin{figure}[h]
\centerline{\includegraphics[width=1.0\columnwidth]{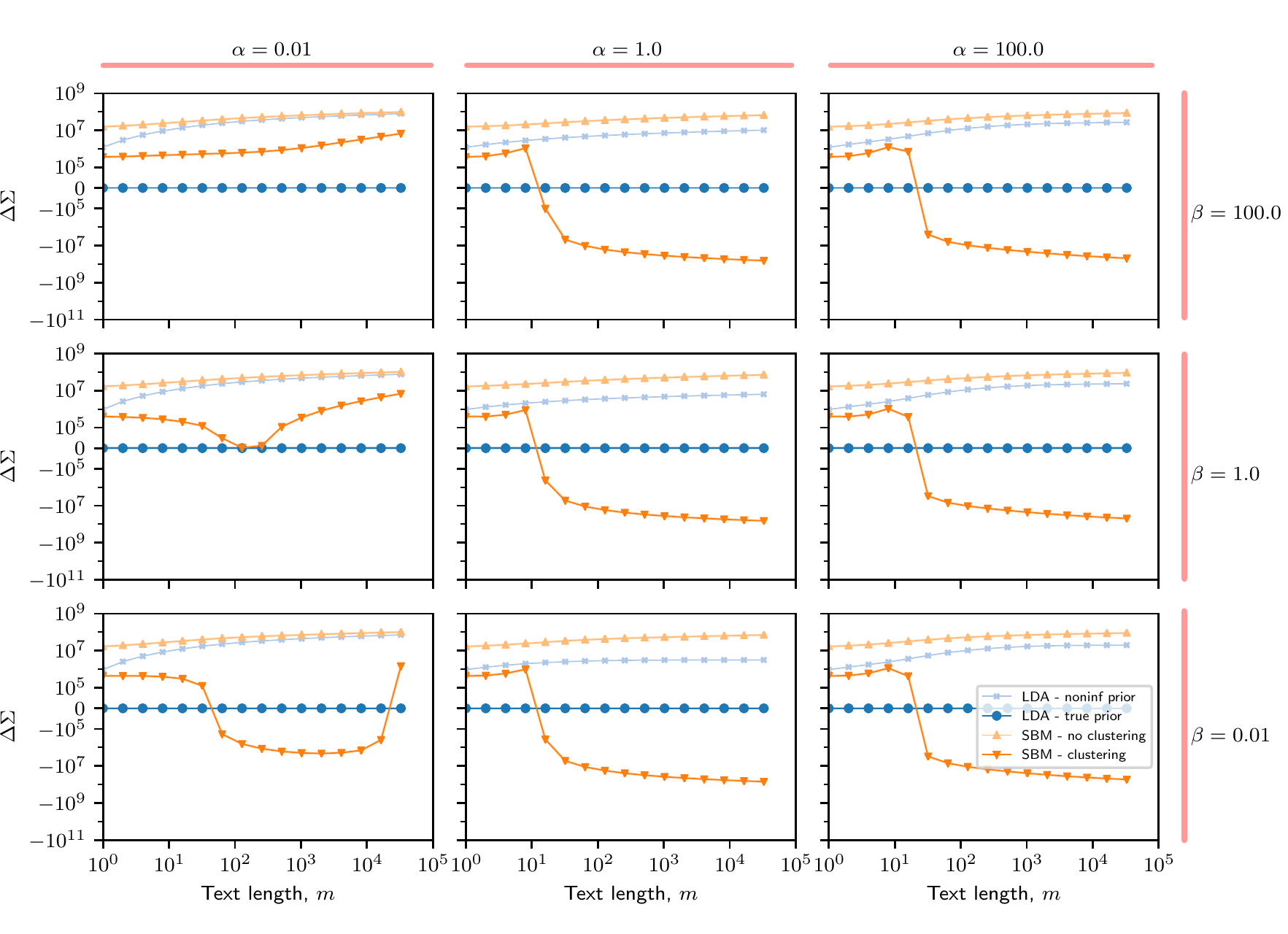}}
\caption{Varying the hyperparameters $\alpha$ and $\beta$ in the comparison between LDA and SBM for artificial corpora drawn from LDA.
Same as in Fig.~4A (main text) with different values $\alpha \in \{ 0.01,1.0,100.0\}$ and $\beta \in \{ 0.01,1.0,100.0\}$.
Note that the panel in the middle corresponds to Fig.~4 in the main text. }
\label{fig.drawLDA.alphabeta}
\end{figure}

\begin{figure}[h]
\centerline{\includegraphics[width=1.0\columnwidth]{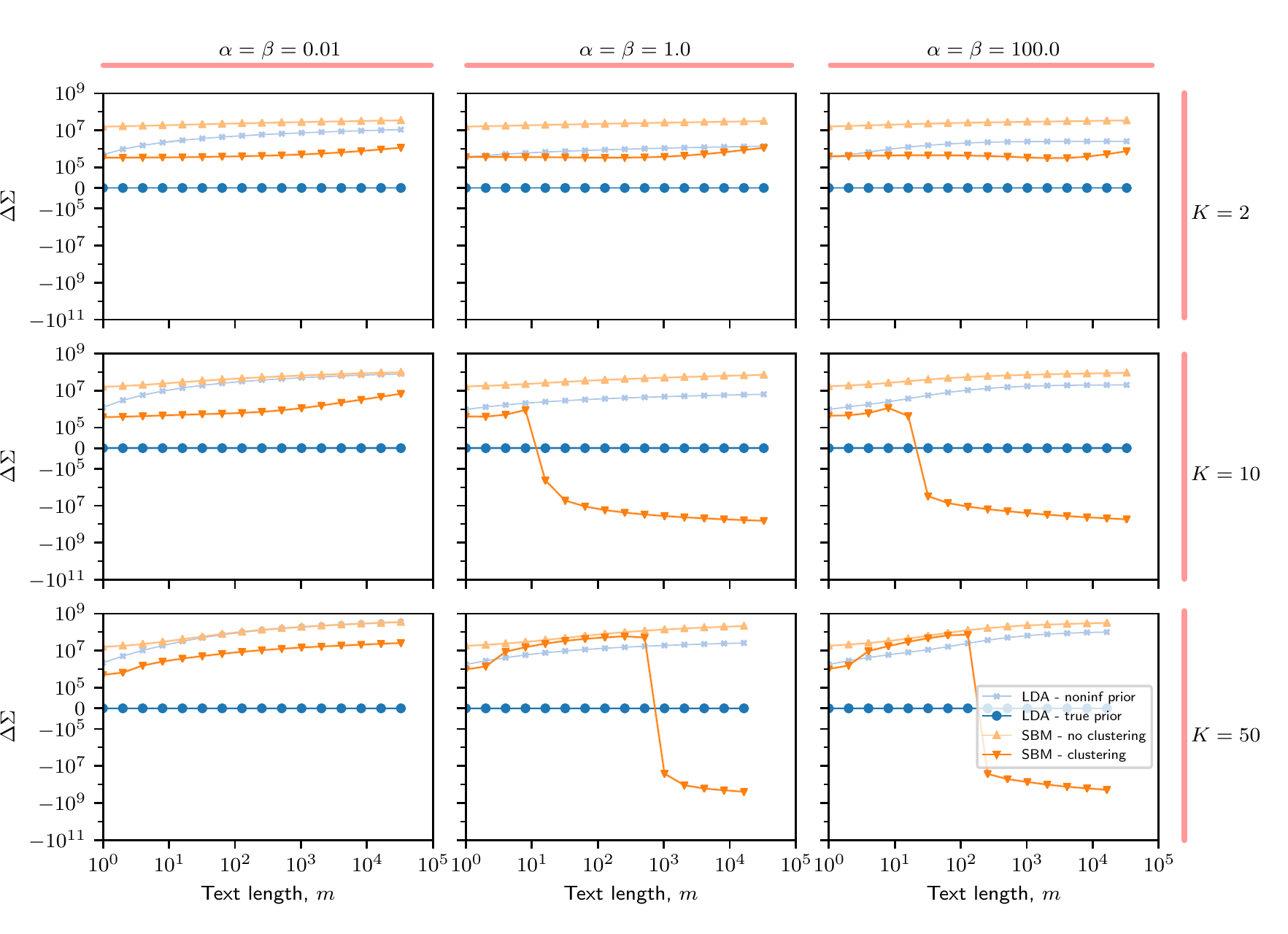}}
\caption{Varying the number of topics $K$ in the comparison between LDA and SBM for artificial corpora drawn from LDA.
Same as in Fig.~4A (main text) with different values $K \in \{ 2,10,100\}$ and $(\alpha,\beta) \in \{ (0.01,0.01),(1.0,1.0),(100.0,100.0)\}$.
Note that the panel in the middle corresponds to Fig.~4 in the main text. }
\label{fig.drawLDA.alphaK}
\end{figure}

\begin{figure}[h]
\centerline{\includegraphics[width=1.0\columnwidth]{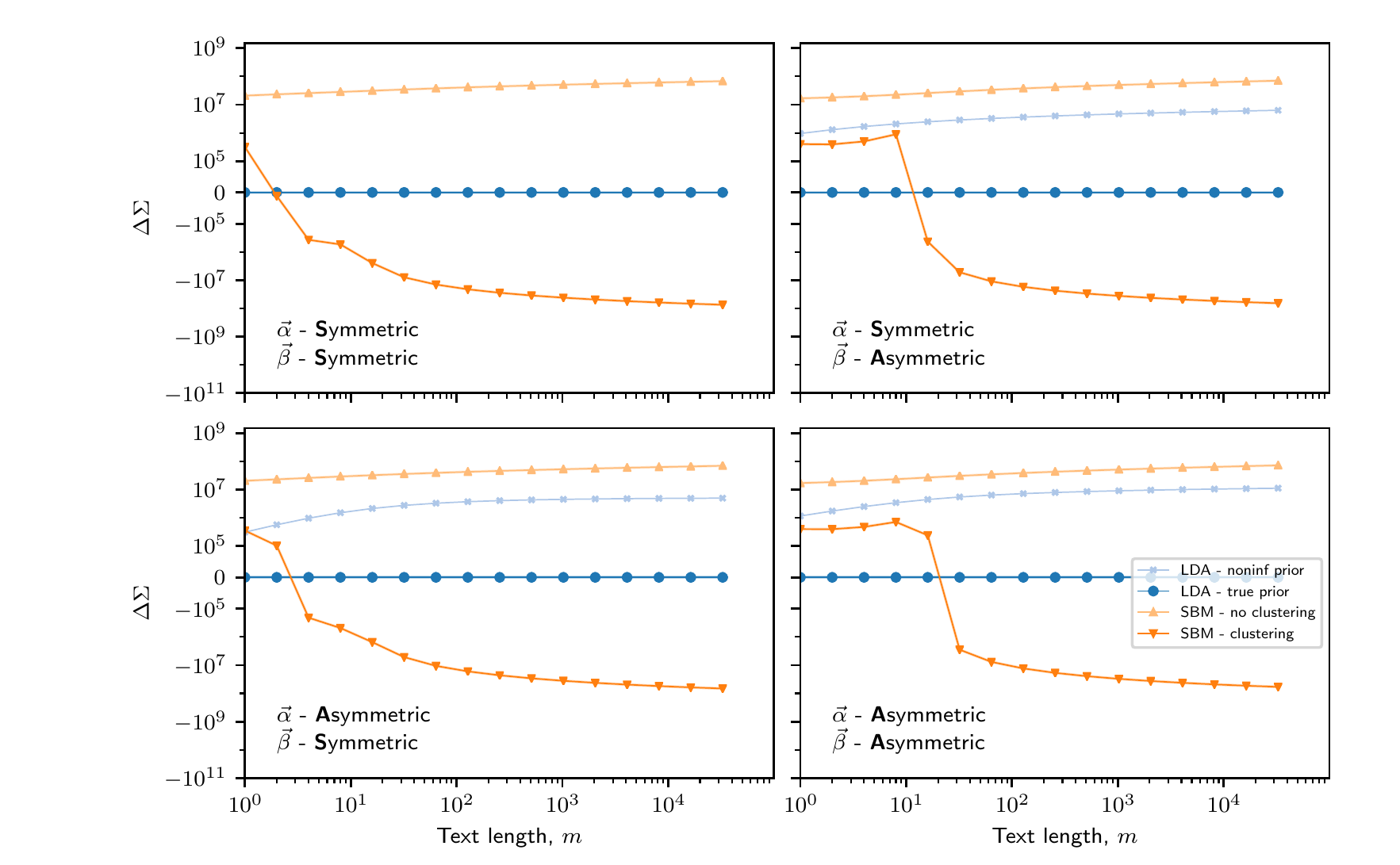}}
\caption{
Varying the base measure of the hyperparameters $\vec{\alpha}$ and $\vec{\beta}$ in the comparison between LDA and SBM for artificial corpora drawn from LDA.
Same as in Fig.~4A (main text) with different symmetric and asymmetric $\vec{\alpha}$ and $\vec{\beta}$.
For $\vec{\alpha}$, the symmetric case is given by $\alpha_{dr}=\alpha$ and the asymmetric case is given by $\alpha_{dr}=\alpha \times K \times p_r $ with $p_r \propto r^{-1}$ for $r=1,\ldots,K$ and $\sum_r p_r = 1$.
For $\vec{\beta}$, the symmetric case is given by $\alpha_{wr}=\beta$ and the asymmetric case is given by $\beta_{wr}=\beta \times V \times p_w $ with $p_w$ empirically measured in Fig.~\ref{fig.rankfrequency.emp} and $V$ is the number of word-types.
}
\label{fig.drawLDA.as}
\end{figure}

\section{Word-document networks are not sparse }
Typically, in community detection it is assumed that networks are sparse, i.e. the number of edges $E$ scales linearly with the number of nodes $N$, i.e. $E \propto N$ ~\cite{fortunato.2010}.
In Fig.~\ref{fig.nodesedges.growth} we observe a different scaling for word-document networks, i.e. a superlinear scaling $E \propto N^{\delta}$ with $\delta>1$.
This is a direct result of the sublinear growth of the number of the number of different words with the total number of words in the presence of heavy-tailed word-frequency distributions (known as Heaps' law in quantitative linguistics~\cite{altmann.book2016}), which leads to the superlinear growth of the number of edges with the number of nodes. 
This means that the density, i.e. the average number of edges per node, increases as more documents are added to the corpus.

\begin{figure}[h]
\centerline{\includegraphics[width=0.5\columnwidth]{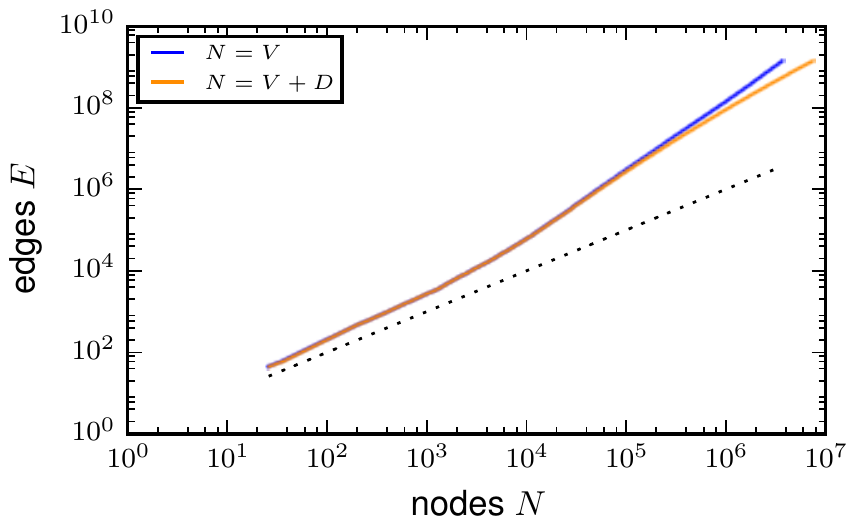}}
\caption{Word-document networks are not sparse. The number of edges, $E$, as a function of the number of nodes, $N$, for the word-document network from the English Wikipedia.
The network is grown by adding articles one after another in a randomly chosen order. 
Shown are the two cases, where i) only the $V$ word-types are counted as nodes ($N=V$) and ii) both the word-types and the documents are counted as nodes ($N=V+D$).
For comparison we show the linear relationship $E=N$ (dotted). 
Figure adapted from Ref.~\cite{gerlach.thesis}.}
\label{fig.nodesedges.growth}
\end{figure}

\section{Empirical word-frequency distribution}
In the comparison of hSBM and LDA for corpora drawn from the generative process of LDA, we parametrize the word-topic hyperparameter as $(\beta_{rw})=(\beta_w) \equiv \bm{\beta}$ for $r=1,\ldots,K$ with $\bm{\beta} = \beta V p_w$ for $w=1,\ldots,V$.
We use an empirical word-frequency distribution $p_w$ as measured from all articles in the Wikipedia corpus contained in the categories ``Scientific Disciplines''.
In Fig.~\ref{fig.rankfrequency.emp} we show the empirically measured rank-frequency distribution for $V=95129$ different words and $M=5,118,442$ word-tokens in total. 
We observe that this distribution is characterized by a heavy-tailed distribution with two power-laws.
In Ref.~\cite{gerlach.2013} it has been shown that virtually any collection of documents follows such a distribution of word frequencies.

\begin{figure}[h]
\centerline{\includegraphics[width=0.5\columnwidth]{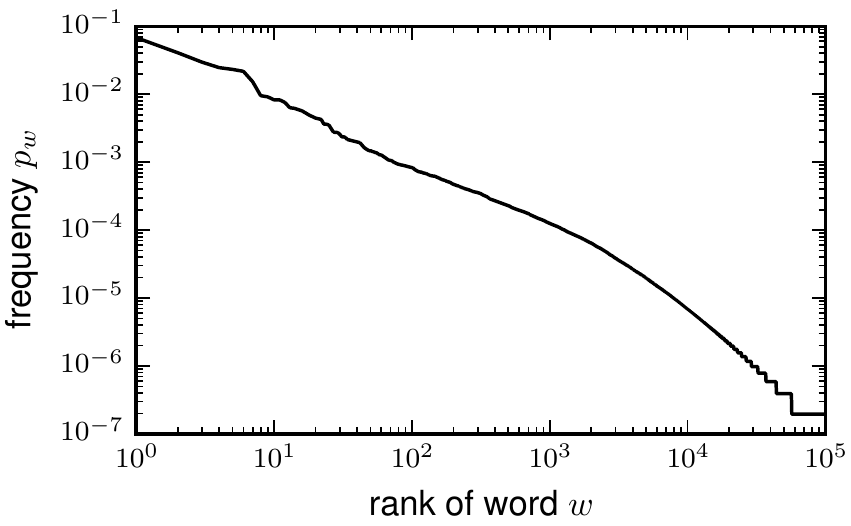}}
\caption{Empirical rank-frequency distribution. The rank-frequency distribution shows the frequency of each word, $p_w=n_w/M$, ordered according to their rank, where $n_w$ is the number of times word $w$ occurs and $M=\sum_w n_w$ is the total number of words. A word is assigned rank $r$ if it is the $r$-th most frequent word, i.e. the most frequent word has rank $1$.
}
\label{fig.rankfrequency.emp}
\end{figure}

\end{widetext}

\end{document}